\documentclass{article}

    \usepackage[preprint]{neurips_2025}

\usepackage[utf8]{inputenc} % allow utf-8 input
\usepackage[T1]{fontenc}    % use 8-bit T1 fonts
\usepackage{hyperref}       % hyperlinks
\usepackage{url}            % simple URL typesetting
\usepackage{booktabs}       % professional-quality tables
\usepackage{amsfonts}       % blackboard math symbols
\usepackage{nicefrac}       % compact symbols for 1/2, etc.
\usepackage{microtype}      % microtypography
\usepackage{xcolor}         % colors
\usepackage{soul}
\usepackage{booktabs}
\usepackage{lineno}
\usepackage{subfigure}
\usepackage{soul}
\usepackage{booktabs}
\usepackage{amsmath}
\usepackage{graphicx}
\usepackage{cleveref}
\usepackage{bm}

\title{Repurposing Marigold for Zero-Shot Metric Depth Estimation via Defocus Blur Cues
}

\newcommand{\Ib}{\mathbf{x}_\text{b}}
\newcommand{\I}{\mathbf{x}}
\newcommand{\Ibhat}{\hat{\mathbf{x}}_\text{b}}
\newcommand{\epsBf}{\mathbf{\varepsilon}}
\newcommand{\zT}{\mathbf{z}_{T}^{(\mathbf{d})}}
\newcommand{\dm}{\mathbf{d}^{\text{m}}}
\DeclareMathOperator*{\argmin}{arg\,min}
\definecolor{myorange}{HTML}{ff9900}
\newcommand{\varcolor}[1]{\textcolor{myorange}{#1}}
\definecolor{myorange}{RGB}{255,140,0}  % or use HTML like {HTML}{FF8C00}
\newcommand{\orange}[1]{\textcolor{myorange}{#1}}
\author{%
  Chinmay Talegaonkar\thanks{Corresponding author: \texttt{ctalegaonkar@ucsd.edu}}, 
  Nikhil Gandudi Suresh, 
  Zachary Novack, 
  Yash Belhe, \\
  \textbf{Priyanka Nagasamudra, 
  Nicholas Antipa} \\
  University of California San Diego
}
\begin{document}

\maketitle

\begin{abstract}
Recent monocular metric depth estimation (MMDE) methods have made notable progress towards zero-shot generalization. However, they still exhibit a significant performance drop on out-of-distribution datasets. We address this limitation by injecting defocus blur cues at inference time into Marigold, a \textit{pre-trained} diffusion model for zero-shot, scale-invariant monocular depth estimation (MDE). Our method effectively turns Marigold into a metric depth predictor in a training-free manner.
To incorporate defocus cues, we capture two images with a small and a large aperture from the same viewpoint. To recover metric depth, we then optimize the metric depth scaling parameters and the noise latents of Marigold at inference time using gradients from a loss function based on the defocus-blur image formation model. We compare our method against existing state-of-the-art zero-shot MMDE methods on a self-collected real dataset, showing quantitative and qualitative improvements. 
\end{abstract}

\section{Introduction}
Estimating metric depth from a single camera viewpoint is a central problem in computer vision with numerous downstream applications, including
3D reconstruction \cite{jiang2024construct}, autonomous driving \cite{schon2021mgnet}, and endoscopy \cite{liu2023self}. 
This task, known as \textit{monocular metric depth estimation} (MMDE), is fundamentally ill-posed due to inherent depth-scale ambiguity \cite{saxena2005learning}. 
Multi-view methods \cite{wen2025foundationstereo} avoid this ambiguity but are often expensive and impractical in settings like endoscopy or microscopy. 
Training data-driven MMDE methods is challenging, as it requires accounting for a diverse set of camera parameters and metric depth scales. 
As a result, existing MMDE models struggle in zero-shot settings, i.e., they generalize poorly to unseen datasets. 
Recent advances in zero-shot (MMDE)~\cite{zhang2025surveymonocularmetricdepth} have demonstrated improved generalization, but there is still a considerable performance drop on unseen datasets. 
% MMDE is fundamentally ill-posed, since all points along a camera ray map to the same pixel in the image. 

% Recent advances in transformer-based foundation models and internet-scale datasets have led to immense progress in zero-shot monocular metric depth estimation (MMDE) \cite{zhang2025surveymonocularmetricdepth}, with strong in-the-wild generalization. 
% MMDE is central to many applications in computer vision, robotics, autonomous driving, AR/VR, microscopy, endoscopy, and computational imaging \ct{cite}. MMDE is fundamentally ill-posed because any point along a (pinhole) camera ray projects to the same pixel. 
% While stereo and multi-view camera setups do not have such an ambiguity, these setups are typically larger, costlier, and may not be easily feasible in settings such as endoscopy or microscopy.  \ct{check with Nick how to write this better}. 

In contrast to MMDE, monocular \textit{relative} depth estimation (MDE) methods recover a relative depth map, factoring out the physical depth scale. 
This enables using large-scale datasets with diverse depth ranges \cite{ranftl2020towards} for training data-driven MDE methods. As a result, MDE methods achieve better zero-shot generalization at significantly lower training cost than MMDE methods, as shown by recent results~\cite{yang2024depth, yang2024depthv2, ke2024repurposing}.
However, despite favorable performance on benchmarks, the absence of metric scale in MDE outputs precludes their applicability in downstream tasks requiring absolute depth. 

Existing data-driven MMDE methods commonly suffer from two failure modes: undesirable coupling between image texture and depth predictions (\cref{fig:synthetic_texture}), and inaccurate estimation of the scene’s physical scale (see \cref{fig:realresults}). The first issue of \textit{texture coupling} also affects MDE methods (see \cref{fig:synthetic_texture}) unless explicitly mitigated through complex training procedures that account for texture variation~\cite{he2024lotus, song2025depthmaster}. 

We demonstrate that incorporating camera physics, particularly defocus blur, with a data-driven MDE model at inference time can effectively address both failure modes \textit {without any re-training}. We use Marigold \cite{ke2024repurposing}, a diffusion-based MDE model,
as it enables backpropagation-based inference time optimization using defocus cues, while staying on its learned manifold of plausible depth maps. To provide these defocus blur cues, we capture two images from a single viewpoint: a small-aperture all-in-focus (AIF) image, which is given as input to Marigold, and a large-aperture image that provides physical cues for metric depth. Our approach requires only a variable aperture camera, such as a DSLR, and knowledge of the lens focal length, focus distance, and F-stop, which are readily available from image metadata; this avoids the need for extrinsic calibration required by multiview setups.

% Requiring only a variable aperture camera, our approach is compatible with standard off-the-shelf DSLR cameras. We only assume access to the focal length and F-stop (available in image metadata), obviating extrinsic camera calibration typically needed in multiview setups.  
% In contrast, transformer-based methods \cite{yang2024depth, yang2024depthv2} would require retraining to adapt to new data \cite{yeo2023rapid}.

% While transformer-based MDE methods \cite{yang2024depth, yang2024depthv2} are faster, they require re-training/finetuning \cite{yeo2023rapid} for adapting to different data distributions. We prefer a diffusion-based MDE method \cite{ke2024repurposing} as it allows integrating physical cues and better controllability at inference time without any re-training. 
% We leverage optical defocus cues in conjunction with a monocular depth estimator at inference time for metric depth estimation. 
% While transformer-based MDE methods \cite{yang2024depth, yang2024depthv2} are faster, they require re-training/finetuning \cite{yeo2023rapid} for adapting to different data distributions. We prefer a diffusion-based MDE method \cite{ke2024repurposing} as it allows integrating physical cues and better controllability at inference time without \textit{any} re-training. 
\begin{figure}
\centering
\includegraphics[width=\linewidth]{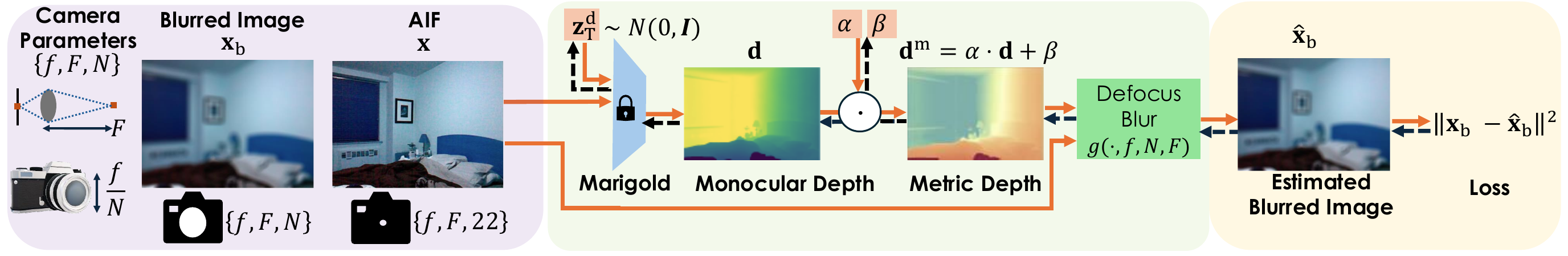}
\caption{\textbf{Method overview.} We capture two images (same viewpoint) from a camera with focal length $f$ and focused at a distance $F$: an all-in-focus (AIF) image $\I$ (F-stop: $N=22$) and a blurred image $\Ib$ (F-stop: $N<22$). Using the AIF $\I$ and an initial \textit{learnable} noise vector $\zT$, Marigold predicts the relative depth $\mathbf{d}$. We then affine transform $\mathbf{d}$ with \textit{learnable} parameters ($\alpha,\beta$), obtaining the metric depth $\mathbf{d}^{\text{m}}$. Given the AIF $\I$, depth $\mathbf{d}^{\text{m}}$, and camera parameters ($f, F, N$), we synthesize a blurred image $\Ibhat$ using the defocus blur forward model. To update the learnable parameters, we compute their gradients w.r.t the L2 loss between $\Ibhat$ and $\Ib$.}
% \na{use cdot for multiplication in the figure (above metric depth). Also, should it be $\alpha$ and $\beta$ for the affine parameters in the figure? Or $a'$ and $b'$? For the 2-norm, if typesetting in Powerpoint, which you really shouldn't be doing ;), make sure to use the proper norm function so that $\|I_b - \hat{I}_b\|$ doesn't have two different sized bars.  Make sure any equations, functions, or variables used here are labeled in a way that corresponds to the equations in the main text. For example, the Defocus Blur has an equation, $g(\cdot)$ which you use later. Go ahead and include in the box. Shouldn't the blurred image and AIF also output $f $ and $N$, which are input to Defocus Blur box?}
\label{fig:method}
\end{figure}

% The small aperture (blur-free) image serves as the input to an MDE, while the large aperture image provides physical cues for metric depth}. 
% Using a single camera viewpoint removes the need for calibrating camera extrinsics -- which is typically needed for multiview/stereo depth approaches.

Defocus blur is known to provide coarse metric depth cues in monocular settings \cite{maximov2020focus, tang2017depth, xiong1993depth}, but has rarely been explored alongside recent data-driven advances in MMDE. We show in our work that revisiting this classical cue and integrating it with modern relative depth methods like Marigold improves metric depth estimation performance, without any retraining. While most depth from defocus methods use multiple measurements~\cite{hazirbas2019deep, liu2017matting, wang2021bridging}, we require only two. Prior work has also explored other hardware novelties to encode depth cues, such as dual pixel cameras \cite{ghanekar2024passive, xin2021defocus} and coded apertures \cite{levin2007image}. However, these depth cues are coarse, and training MMDE models for these systems is limited by the lack of datasets captured with a sufficient camera parameter diversity. This motivates the need for an approach that combines strong data-driven pre-trained MDE/MMDE models trained on existing RGB datasets with physics-based cues from non-pinhole cameras.

To this end, our \textbf{contributions} are as follows: We formulate MMDE as an inverse problem under the defocus blur image formation model. Using inference-time optimization, we recover the metric depth scale and correct for texture-depth coupling -- without retraining Marigold.
We devise a hardware setup consisting of a rigidly coupled DSLR (RGB camera) and an Intel RealSense depth camera to capture RGB images and the ground truth metric depth. We collect a dataset of 7 diverse real-world indoor scenes, captured at different defocus blur levels. We compare our method on this dataset with current MMDE methods and demonstrate quantitative improvements on real data. 
\section{Related Work}
\label{sec:relatedwork}
\paragraph{Zero-shot depth models}
Recent approaches for Monocular (relative) Depth Estimation (MDE) and Monocular Metric Depth Estimation (MMDE) can be classified as discriminative or generative. Discriminative approaches largely rely on vision transformer-based architectures \cite{bochkovskii2024depth, bhat2023zoedepth, birkl2023midas, yin2023metric3d, piccinelli2024unidepth} trained on large-scale datasets. Transformer-based methods have been more successful for MDE \cite{yang2024depth, yang2024depthv2} compared to MMDE accuracy-wise, as MMDE is a more ill-posed task than MDE. These approaches have very low inference times, but for MMDE, the performance degrades on out-of-distribution test scenes\footnote{Refer to \cite{zhang2025surveymonocularmetricdepth} for a detailed survey on MMDE methods.}. Most of these methods are purely data-driven and ignore depth-based visual effects in images such as defocus cues. It is also difficult to incorporate physics-based depth refinement in these methods at test time without any re-training \cite{yeo2023rapid}. Since we use a generative MDE, our approach can incorporate physical cues at test time in a training-free manner. 

Current state-of-the-art \textit{generative} MDE/MMDE methods \cite{saxena2023diffusion, duan2024diffusiondepth} are predominantly diffusion-based. Several previous methods \cite{duan2024diffusiondepth, saxena2023diffusion, liu2021swin} incorporate diffusion-based depth denoising in their pipelines, achieving highly detailed depth maps -- but are not zero-shot. \cite{saxena2023zero} achieves zero-shot MMDE with a diffusion-based approach by incorporating diverse field of view (FOV) augmentations in training, but it is not open source and lags behind transformer-based methods in performance.
Marigold~\cite{ke2024repurposing} is trained by fine-tuning Stable Diffusion-v2 on synthetic depth data. It achieves high-quality zero-shot MDE and supports test time refinement, but is not applicable natively for MMDE. Our approach uses defocus cues to refine the relative depth predictions from Marigold (or similar methods \cite{fu2024geowizard}), enabling its application to MMDE. Prior work \cite{viola2024marigold} also proposes a similar strategy for dense MMDE from a sparse metric depth map using Marigold and test time optimization. In contrast, our method does not require a sparse depth map as input, and solely relies on RGB images and a priori known scene bounds. Using defocus blur cues, our method resolves inaccuracies in monocular depth while also estimating the global scaling parameters for metric depth. 
% \ct{The last sentence should be more reminiscent of the thesis in the intro.}
% \begin{itemize}
%     \item These works do not use lens blur and other knowledge of the physical camera system.
%     \item They are trained based on common depths of different objects, but it is infeasible to get accurate metric depth from this as it is a very ill-posed problem
%     \item Due to this, the metric depth estimates are not very accurate as shown in our studies
%     \item This includes ZoeDepth, Depth Pro, Zero-Shot Metric Depth, DiffusionDepth...
%     \item For relative depth includes Marigold, ...
% \end{itemize}
\paragraph{Diffusion model priors for inverse problems}
We frame MMDE as an inverse problem under the defocus blur image formation model. This framing closely relates to the recent work on solving linear inverse problems using \textit{pre-trained} diffusion models \cite{daras2024survey}. \cite{chung2023diffusion, chung2022come, song2023pseudoinverse, chung2022improving} incorporate the forward model constraints while sampling pixel-space diffusion models pre-trained on smaller datasets. As a result, these methods require many steps while sampling the diffusion model and have limited generalizability as priors. \cite{song2023solving, rout2024solving} use latent variable diffusion models (LDMs) as priors, resulting in better in-the-wild generalizability. We use Marigold as the LDM, but instead of incorporating the defocus forward model during sampling, we optimize the latent noise vector based on the error between the observed and predicted image using the forward model. Our approach is inspired by recent methods \cite{novack2024ditto, Novack2024Ditto2, Zheng2023TiNOEdit, Wallace2023EndtoEndDL, Gal2022RareConcepts}  
which uses noise optimization in conjunction with a differentiable auxiliary guidance loss to improve the sampling quality of the diffusion model based on text input. While these methods use trained models as differentiable proxies for guidance, we use a physics-based imaging forward model. \cite{nathan2024osmosis} also uses a physics-based forward model with a diffusion prior, but requires re-training the diffusion prior from scratch. 
% \ct{find a way to clarify what generalization capability as prior means}. 
% Refer to \cite{daras2024survey} for details on more works in this space.
% \ct{add a sentence that says }

% \begin{itemize}
%     \item Most work seems to be done in text-to-image space where more accurate images to represent the text is done.
%     \item There are many methods of doing noise vector optimization. For example using attention layers to divide noise vector space.
%     \item Many noise vector optimization do not deal with conditional image diffusion models, although the same principles applies
%     \item (Hopefully) We use chi-square optimization to update noise vector as in Samuel et all.
% \end{itemize}

% \subsection{Diffusion Models for inverse problems}
% \begin{itemize}
%     \item Many works assume a well-known forward model
%     \item For us, depth is a very nonlinear problem for lens blur and highly subject to noise
%     \item A lot of these forward models are also a lot smaller, it is hard to directly invert ours without gradient methods.
%     \item Most of them assume spatially invariant blur kernels. We don't. 
% \end{itemize}

\paragraph{Depth estimation from passive camera physics cues}
A substantial body of research on MMDE leverages camera physics, including methods like depth-from-defocus (DfD),
-\cite{watanabe1998rational, gur2019single, subbarao1994depth, alexander2016focal}, phase/aperture masks \cite{levin2007image, zhou2009coded, zheng2020joint, antipa2017diffusercam, wu2019phasecam3d}, and dual pixel sensors \cite{garg2019learning, xin2021defocus, abuolaim2020defocus}. Classical approaches produce very coarse depth maps. Most DfD methods need a well-aligned \cite{won2022learning} multi-image focal stack \cite{lin2015depth, strecke2017accurate}, to achieve good depth quality. Optical mask-based methods pose a harder inverse problem of jointly estimating both AIF and the depth map. We capture only 2 images at the same focus distance and different apertures, requiring no image alignment and AIF estimation. This simplifies depth refinement with minimal added capture time. Previous work
has also used variable apertures for classical \cite{farid1998range} and learning-based \cite{srinivasan2018aperture} depth estimation methods, but it is not zero-shot. While learning-based approaches \cite{ghanekar2024passive, carvalho2018deep, won2022learning, garg2019learning} help improve the depth map quality for these methods, they don't generalize well to out-of-distribution scenes. Dual pixel-based methods are popular for phone cameras,  \cite{garg2019learning, pan2021dual}, but are tied to the specific camera architecture. While we show results with a standard DSLR sensor, our approach can be adapted to dual-pixel-style camera architectures as well. 
% \begin{itemize}
%     \item These works either require some sort of training or they do not take advantage of better priors.
%     \item Training works require training which is a big downside. 
%     \item They cannot generalize to new images as well
%     \item Unidepth, dual pixel, Srinivasan et all  ... need to find more in this category actually
%     \item The non-training ones like Levin et all, Flat Cam, ... cannot take advantage of more modern diffusion priors.
% \end{itemize}

% \subsection{Dual Pixel and Coded Aperture Stuff}

% ---------

% \begin{itemize}
%     \item Our method uses accurate inference time optimization to make diffusion model input correspond with lens blur models
% \end{itemize}

\section{Preliminaries}
\label{sec:prelims}
\paragraph{Diffusion-based monocular depth estimation} 
%%% make this training paragraph
% \ct{Maybe add a few lines on training marigold here.}
Our approach is built on top of Marigold \cite{ke2024repurposing}, which is a monocular depth estimator trained by fine-tuning the denoising U-Net 
% $\epsilon_{\theta}(.)$
of StableDiffusion-v2 (SDv2) \cite{rombach2022high} on synthetic depth data. Marigold allows sampling the conditional distribution of the monocular depth given an input image,  $p(\mathbf{d}_0|\mathbf{x})$. 
% \na{Should this be $\mathbf{I}$ instead of $\mathbf{x}$, or am I misunderstanding later notation?} % \na{doesn't it only implicitly approximate it by learning a generative function that provides sampling from it? It doesn't actually provide p(d|x)}. 
In particular, Marigold attempts to generate clean monocular depth maps, $\mathbf{d}_0$ given a clean input image $\mathbf{x}$ by first sampling $\mathbf{d}_T \approx \mathcal{N}(0, \bm{I})$ from an i.i.d Gaussian distribution, and then iteratively denoising it (where each intermediate step is denoted by $\mathbf{d}_t$) according to a fixed noise schedule with parameters $\alpha_t, \sigma_t$.
% to progressively cleaner depth estimates $\mathbf{d}_t$; starting with 
% where $\mathbf{d}_T \approx \mathcal{N}(0, \bm{I})$, and the corruption process is determined by a priori chosen noise schedule $\alpha_t, \sigma_t$)
As SDv2 is a \emph{latent}-diffusion model, the entire generative process happens on encoded latent depth maps $\mathbf{z}_{0}^{\mathbf{(d)}}$ with latent images $\mathbf{z}^{\mathbf{(x)}}$ as conditioning. Training Marigold\footnote{While technically SDv2 is trained in $\epsilon$-prediction mode rather than $\mathbf{x}$-prediction, we use $\mathbf{x}$-prediction given their mathematical equivalence and alignment with Marigold-LCM, which uses $\mathbf{x}$-prediction.} (which we denote as $\hat{\mathbf{x}}_{\bm\phi}(\cdot)$) involves using the standard denoising loss common in image diffusion works, but on depth maps rather than images, and passes the image input into the model as well through channel-wise concatenation:
\begin{align}
\mathbb{E}_{\mathbf{z}_{0}^{\mathbf{(d)}}, \mathbf{z}^{\mathbf{(x)}} \sim \mathcal{D}, t \sim p(t), \epsilon \sim \mathcal{N}(0, \bm{I})}[w(t)\|\mathbf{z}_{0}^{\mathbf{(d)}} - \hat{\mathbf{x}}_{\bm\phi}(\alpha_t\mathbf{z}_{0}^{\mathbf{(d)}} + \sigma_t\epsilon,\mathbf{z}^{\mathbf{(x)}}, t)\|_2^2],
\end{align}
where $w(t)$ is a time-dependent weighting function. At inference time, $\mathbf{x}$ is first mapped to a lower dimensional latent vector $\mathbf{z^{(x)}} = \mathcal{E}(\mathbf{x})$ through VAE encoder $\mathcal{E}$ in SDv2. The inference process starts with sampling a noisy latent depth vector $\mathbf{z}_{T}^{\mathbf{(d)}} \sim \mathcal{N}(0, \bm{I})$, which is iteratively refined by applying the denoiser $\hat{\mathbf{x}}_{\bm\phi}(\mathbf{z}_{t}^{\mathbf{(d)}}, \mathbf{z^{(x)}}, t)$ to obtain $\mathbf{z}_{0}^{\mathbf{(d)}}$ over $T$ sampling steps. $\mathbf{z}_{0}^{\mathbf{(d)}}$ is then decoded through the SD decoder $\mathcal{D}$ to produce the output depth map $\mathbf{d} = \mathcal{D}(\mathbf{z}_{0}^{\mathbf{(d)}})$. Marigold assumes both $\mathbf{z}^{(\mathbf{x})}, \mathbf{z}^{(\mathbf{d})} \in \mathbb{R}^{M}$. For faster inference, we use the latent consistency model version of Marigold, Marigold-LCM, and obtain $\mathbf{z}_{0}^{\mathbf{(d)}}$ with a single inference step from $\zT$. Monocular depth  $\mathbf{d}$ can be mapped to metric depth $\mathbf{d}^{\text{m}}$ by an affine transform; more details in \cref{sec:method}.
% \ct{Add more details on Marigold training.}
% \ct{More lucidly introduce how to go from diffusion to GANs}
% \begin{align}
% \label{eq:affine_depth_transform}
%     \mathbf{d}^{m} = a.\mathbf{d} + b
% \end{align}

% \ct{Nick -- the modeling defocus cues paragraph}

\begin{figure}
    \centering
    \includegraphics[width=0.9\linewidth]{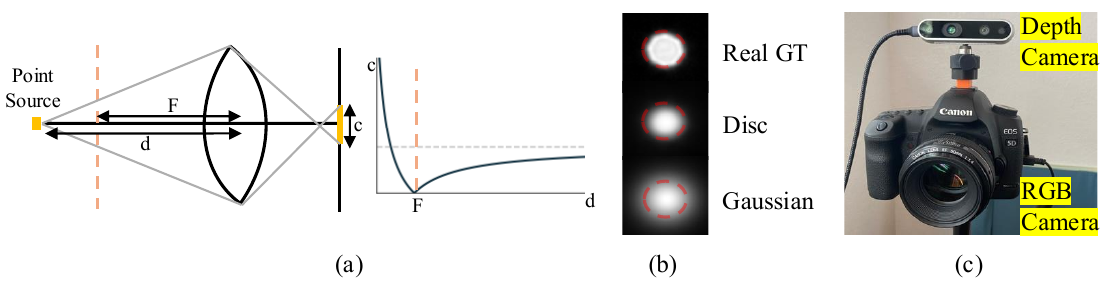}
    \caption{\textbf{Comparing simulated PSFs with the PSF captured from our camera setup.} (a) A point source placed $d$ distance away from a thin lens focused at a focus distance $F$ produces a blurred image (PSF) with a diameter $c$, also known as the circle of confusion. The variation of $c$ with source distance $d$ is shown in the plot. (b) The Disc approximation to the camera PSF lies roughly within the same bounds (dotted red circle) as the PSF captured from the RGB camera (Real GT) in (c). The Gaussian PSF significantly exceeds the bounds. Slight differences between the real and Disc PSF stem from the octagonal aperture and diffraction ignored in our model. (c) We rigidly mount an Intel RealSense on a DSLR to capture ground truth depth, and calibrate both cameras to align predicted depth from the RGB image with the ground truth depth for evaluation.
    % We rigidly mount an Intel RealSense (depth camera) on a DSLR camera to capture the ground truth depth. We calibrate both the cameras to align the ground truth (realsense) and predicted  (DSLR camera) depth maps for evaluation.
    % The slight differences in the real and Disc PSF arise from the octagonal aperture shape and diffraction effects, which are ignored in our model. 
    % \na{I think fig 2 could be improved: show a surface to the left of the lens which represents the depth map. Then draw rays from two different points, showing the two different PSFs. Label them $h(x-u_i,y-v_i|d(u_i,v_i)$. That captures the PSF model completely in a single figure.}
    }
    \vspace{-1.4em}
    \label{fig:psfcomparison}
\end{figure}
\paragraph{Modeling defocus cues}
Under the thin lens camera assumption, defocus blur manifests as a uniform blur kernel, with a depth-dependent diameter. The blur kernel diameter for a point source placed at a distance $d$ away from the camera is determined by the circle of confusion (CoC)~\cite{10.1145/800224.806818} equation
\begin{align}
\label{eq:coc}
        c(d) = \frac{f^2}{N} \frac{|d-F|}{d(F-f)s},
\end{align}
where the focal length $f$, focus distance 
$F$, and F-stop $N$ are camera parameters that are known from the image EXIF data. $s$ denotes the pixel size in physical units (m). The CoC defines a depth-dependent point spread function (PSF) $h(i,j \mid u,v,d)$ that predicts the response at pixel coordinate $(i,j)$ from a point source at lateral coordinate $(u,v)$ and depth $d$ under defocus blur. Note that $(u, v)$ can represent depth-normalized pixel coordinates under an ideal pinhole projection, allowing us to write $d = \mathbf{d}^{\text{m}}[u,v]$ (valid for general non-volumetric scenes that assume a single depth value per coordinate). 
We assume that the PSF is shift-invariant for a given depth $d$, i.e. $h(i,j \mid u,v,d) = h(i-u, j-v \mid 0, 0, \mathbf{d}^{\text{m}}[u,v]) = h(i-u,j-v \mid \mathbf{d}^{\text{m}}[u,v])$,  which is simply the on-axis PSF, modeled as a disc (assuming circular aperture) with radius given by CoC equations, translated to be centered at $(i,j)$. To ensure smooth optimization, we include a linear fall-off at the boundary of the discontinuous disc kernel, similar to \cite{wang2023implicit}. The PSF can then be expressed as 
\begin{align}
h(i, j \mid u,v,d) &= \frac{ \widehat{W}(i, j \mid u, v, d) }{ \sum_{i,j} \widehat{W}(i, j \mid u, v, d) }, \textrm{where} \\
\widehat{W}(i, j \mid u,v, d) &= 
\begin{cases}
1, & m \leq \frac{c(d) - 1}{2} \\
\frac{c(d) + 1}{2} - m, & \frac{c(d) - 1}{2} < m \leq \frac{c(d) + 1}{2} \\
0, & \frac{c(d) + 1}{2} > m
\end{cases}, \\ m &=  \sqrt{ (i-u)^2 + (j-v)^2}.
\label{eq:disc_coc}
\end{align}
We assume the PSF to be normalized and explicitly account for exposure and energy balancing during image capture and processing (see \cref{sec:method}). 
While the above PSF can also be approximated as an isotropic 2D Gaussian \cite{gur2019single}, we opt for the disc parameterization in \cref{eq:disc_coc} similar to \cite{wang2023implicit, sheng2024dr}, as it better approximates the PSF of the real camera compared to the Gaussian approximation in \cite{gur2019single}, as shown in \cref{fig:psfcomparison}. 
We can see from \cref{eq:coc} that an image captured with a very small aperture (high $N$) would have negligible defocus blur due to very small CoC values. Such an image is referred to as the all-in-focus (AIF) image, $\I$. 
Given the AIF $\I$, the blurred image $\Ib$ can be approximated (neglecting occlusion) as a spatially varying convolution between $h$ and $\I$,
\begin{align}
    \label{eq:defocus_image}
    \Ib(i,j) = \iint\I(u,v) \cdot h(i-u,j-v;\mathbf{d}^{\text{m}}[u,v]) dudv 
\end{align}
% \na{This normalization forces the PSF to have unit sum, which discards the fact that a larger aperture has a PSF with more energy. Think about it this way: if you have a CoC then you open the aperture, the value in the center of the CoC won't change--the new energy gets contributed to the outer edges of the CoC, so its norm would increase.}
For simplicity, we denote the above image formation forward model as
\begin{align}
\label{eq:forward_model}
    \Ib = g(\I, \mathbf{d}^{\text{m}}, f, F, N).
\end{align}
The AIF image $\I$, captured at a high F-stop, serves as the blur-free input for both Marigold and the camera blur model eq.~\eqref{eq:forward_model}, while the low F-stop image $\Ib$ provides defocus cues for MMDE.
% \na{Include one sentence explicitly stating that the AIF captured at high N is what marigold needs, but low N gives us the depth cues we want}
\paragraph{Inference-time optimization}
Without loss of generality, any generative model (GAN, Diffusion or flow-based) $\phi:\mathbb{R}^{M} \rightarrow \mathbb{R}^{N}$ can be construed as a mechanism to map a simple probability distribution, such as an i.i.d. Gaussian distribution, $\epsBf\sim\mathcal{N}(0, \bm{I}) \in \mathbb{R}^{M}$ to a non-linear $N$-dimensional manifold, such as images or audio, through a differentiable generative process $x=\phi(\epsBf)$. Inference time optimization \cite{novack2024ditto} refers to manipulating the generation process by updating the \textit{initial noise} $\mathbf{\epsBf}$ based on gradients from a differentiable loss function $\mathcal{L}(x)$ on the generated sample $x$,
% \na{source? If this is common enough maybe no citation needed, but if a source is known use it}
\begin{align}
    \epsilon\rightarrow \epsilon - \nabla_{\epsilon}\mathcal{L}(x).
\end{align}
% \na{Can you connect this directly to the math from the diffusion-based MDE preliminaries?}
To further ensure that $\epsilon$ still lies close to the Gaussian manifold after the gradient updates, $\epsBf$ can be rescaled to have a L2 norm of $\sqrt{M}$ as in \cite{samuel2024norm}, which is a valid approximation for samples drawn from a high dimensional Gaussian distribution as per the Gaussian annulus theorem \cite{barany2007gaussian}. This holds for most generative models, as the initial noise vectors are typically high-dimensional $(M > 50)$. In our case, $\mathbf{\varepsilon}$ corresponds to the noise latent $\zT$ in Marigold, which we optimize using a loss function (\cref{eq:optimizationobj}) governed by the defocus blur forward model \cref{eq:forward_model}.
% uses a similar constraint on the norm of the noise vector. \na{this sentence feels like it comes out of nowhere?}

% \begin{itemize}
% 1. 
% \end{itemize}
\section{Method}
\label{sec:method}
% \begin{figure}[h]
%     \centering
%     \subfigure[Method Overview]{
%         \includegraphics[width=.64\linewidth]{figures/pipeline_fig.png}
%         \label{fig:pipeline}
%     }\hfill
%     \subfigure[Diffusion model for sampling from plausible depth maps given an all in focus image.]{
%         \includegraphics[width=0.3\linewidth]{figures/diff_model_explainer.png}
%         \label{fig:refinement}
%     }
%     \caption{\textbf{Overview of our depth estimation pipeline.} (a) The overall pipeline illustrating the flow from input images to refined metric depth map. (b) Use of random initializations to choose different plausible depth ma[s]}
%     \label{fig:method_pipeline}
% \end{figure}
We capture two images per scene: $\I$, with F-stop ($N_{\text{aif}} = 22$) and exposure time $t_{\text{aif}}$, serves as the blur-free all-in-focus (AIF) image. A second image, $\Ib$, is captured at a lower F-stop ($N_{\text{b}} = 8$) with exposure time $t_{\text{b}}$, thereby providing strong depth-varying defocus cues. The forward model in \cref{eq:forward_model} assumes radiometrically linear images (no gamma correction or non-linear processing) and energy constancy between the AIF and blurred images. We use raw images to satisfy these assumptions. Since total captured energy scales with exposure time ($t$) and aperture area ($\propto (f/N)^2$) \cite{jones1926relation}, we scale $\Ib$ by the factor $\frac{t_{\text{aif}}}{t_{\text{b}}} \cdot \frac{N^2_\text{b}}{N^2_{\text{aif}}}$ to match the energy in $\I$. Note that we vary the exposure time to ensure well-exposed measurements across F-stop settings, while fixing the camera gain. 

\begin{figure}
    \centering
    \includegraphics[width=\linewidth]{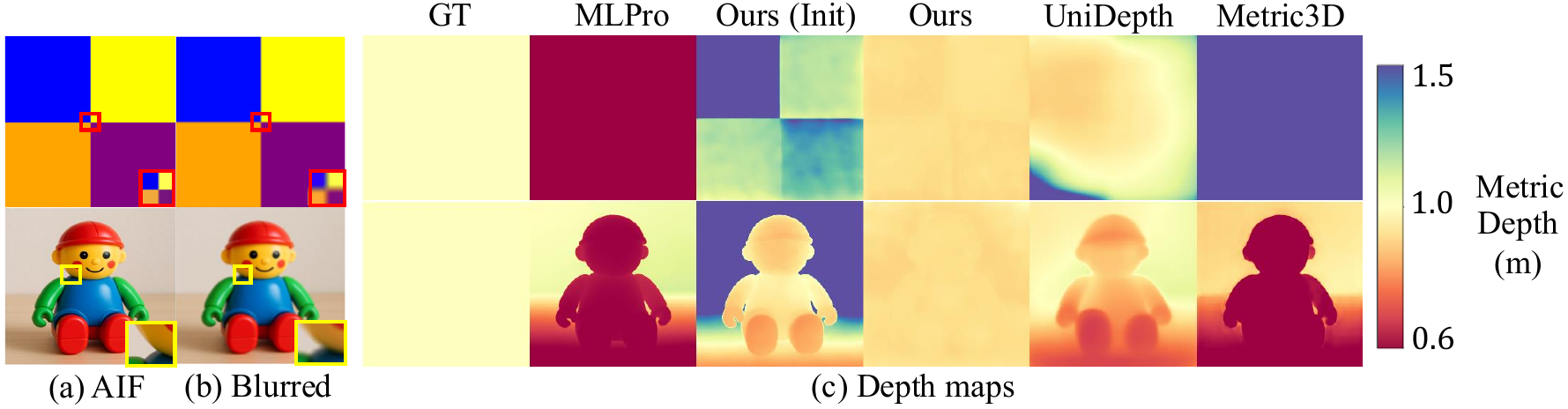}
    \caption{\textbf{Correcting texture-depth coupling.} We assess MMDE performance on textured fronto-parallel 2D planes with constant ground truth depths (GT).  Using an all-in-focus (a) and blurred image (zoom into insets) (b), our method (RMSE: 0.01) recovers the correct depth maps (c) for the two textured planes. We resolve the texture coupling in the Marigold prediction (Ours Init) and recover the correct metric scale. 
    Competing methods (RMSE: 0.2-0.5) fail to predict both the constant relative depth map (except MLPro and Metric3D in row 1) and the correct scale. }
    % \ct{fix this sentence} While MLPro also disambiguates the texture coupling in row 1, for the toy figurine (row 2) it fails to predict the correct depth scale. 
    % The depth maps are all plotted with the same colorbar limits for visual consistency. Our method (RMSE: 0.01) outperforms the baselines (RMSE: 0.2-0.5) by a significant margin. \na{color bar for depth maps}} 
    \label{fig:synthetic_texture}
\end{figure}

We frame metric depth estimation as an inverse problem, with the defocus blur image formation process in \cref{eq:defocus_image} as the forward model, and Marigold as the monocular depth prior. 
To obtain scale-invariant monocular depth $\mathbf{d} \in [0,1]$, we use Marigold-LCM, which takes in as input the AIF $\I$ (encoded to $\mathbf{z}^{(\mathbf{x})}$), and a learnable depth latent vector $\mathbf{z}_{T}^{(\mathbf{d})}\sim \mathcal{N}(0, \bm{I})$.
% \begin{align}
%     \mathbf{z}^{(\mathbf{d})}_{0} = \hat{\mathbf{x}}_{\phi}\left(\zT, \mathbf{z^{(x)}},1\right)
% \end{align}
A single inference step of Marigold-LCM gives us the denoised depth latent $\mathbf{z}_{0}^{(\mathbf{d})} = \hat{\mathbf{x}}_{\phi}\left(\zT, \mathbf{z^{(x)}},1\right)$, which is decoded to monocular depth $\mathbf{d} = \mathcal{D}(\mathbf{z}^{(\mathbf{d})}_0)$. The predicted monocular depth $\mathbf{d} \in [0,1]$ is then mapped to metric depth by affine transforming $\mathbf{d}$ with a learnable metric scale ($\alpha$) and offset ($\beta$) per scene, $\dm = \alpha \cdot \mathbf{d} + \beta$. To ensure that $\alpha, \beta$ remain bounded and differentiable, we parameterize them as $\alpha = s_{\text{max}} \cdot \sigma(a)$ and $\beta = s_{\text{min}} \cdot \sigma(b)$, where $\sigma(\cdot)$ is the sigmoid function; $a,b$ are unconstrained learnable parameters initialized to 0, and $s_{\text{max}}$ and $s_{\text{min}}$ are the upper and lower scene depth bounds, respectively, which we assume are known a priori (valid for indoor scenes). To summarize, the metric depth ($\mathbf{d}^{\text{m}}$) can be expressed using the optimizable parameters $\varcolor{a},\varcolor{b},\varcolor{\zT}$ as: 
\begin{align}
    \mathbf{d}^{\text{m}} &= s_{\text{max}}\cdot \sigma(\varcolor{a}) \cdot \mathcal{D}\left(\hat{\mathbf{x}}_{\phi}\left(\varcolor{\zT}, \mathbf{z^{(x)}},1\right)\right) + s_{\text{min}}\cdot\sigma(\varcolor{b})\nonumber \\
    &:= y\left(a,b, \zT\right) {.}
    \label{eq:metric_depth}
\end{align}
% where $\alpha,\beta$ are the affine scale and offset estimated per scene. 
% \begin{equation}
% \begin{aligned}
%     a = s_{\text{max}} \cdot \sigma(a'),
%     b = s_{\text{min}} \cdot \sigma(b'), 
% \end{aligned}
% \label{eq:sigmoid_param}
% \end{equation}
The optimized metric depth $\widehat{\mathbf{d}}^{\text{m}}$ can then be recovered by solving:
\begin{align} 
    \widehat{\mathbf{d}}^{m} =\argmin_{\mathbf{d}^{\text{m}} = y\left(\varcolor{a},\varcolor{b},\varcolor{\zT}\right)} &\quad || \Ib - g(\I, \mathbf{d}^{\text{m}}, f,F, N)||_{2}^2  \\
    \text{subject to} &\quad \left\|\zT\right\|_2 = \sqrt{M},
    \label{eq:optimizationobj}
\end{align}
where $g(\cdot)$ denotes the defocus blur forward model (eq.~\eqref{eq:forward_model}), $\Ib$ and $\I$ are the captured blurred and AIF images, respectively. By optimizing the learnable parameters $\varcolor{a},\varcolor{b},\varcolor{\zT}$, we incorporate defocus blur cues for both correct metric scale recovery ($a,b$) and refining the initial depth estimate by Marigold ($\zT$). See \cref{fig:method} for the overview of our method.
\paragraph{Motivating Synthetic Toy Examples} We demonstrate that our approach resolves texture-depth coupling, accurately recovering metric depth in a synthetic scene with a textured plane at constant depth. While such a plane may seem simple, distinguishing it from a flat 2D image/poster or an actual 3D scene is challenging when viewed from a single viewpoint.
% We show that our approach resolves texture-depth coupling, accurately recovering metric depth in a simple scene with a textured plane at constant depth. A plane at a constant depth seems simple, but Given images captured from a single viewpoint, it is very difficult to know whether they correspond to an actual 3D scene, or another 2D image (or a flat poster).
% \na{Start off by clearly stating the advantages of what we are doing: fixes texture coupling, and provides high accuracy metric depth in real-world usage}
% Given images captured from a single viewpoint, it is very difficult to know whether they correspond to an actual 3D scene, or another 2D image (or a flat poster). 
Data-driven methods are biased towards predicting depths that reflect surface variations even in the absence of true depth changes, i.e., their outputs are strongly \textit{texture coupled}. However, supplementing the AIF image with a simulated blurred image provides defocus cues that help our method disambiguate a flat poster from a 3D scene, as demonstrated in \cref{fig:synthetic_texture} using toy examples of textured planes with constant depth. 
While learning-based methods and initial Marigold outputs suffer from texture coupling and scale errors, our method corrects both, producing accurate, constant-depth maps that outperform all baselines.
% While Learning based methods and the initial depth predicted by Marigold suffer from texture-coupling and incorrect depth-scale estimation, our method can correct for both these artifacts during optimization, producing depth maps with constant depth and correct depth scales -- far superior to competing baselines.
% However, supplementing the AIF with a blurred image provides defocus cues to disambiguate a flat poster from a 3D scene. We show results on a few toy examples of flat (constant depth) planes with different textures in \cref{fig:synthetic_texture} that demonstrate the ability of our method to perform this disambiguation. 
% We assume knowledge of pixel-accurate ground truth depth and no forward model mismatch. 
% \vspace{-4mm}
\paragraph{Accelerating inference} We use the distilled latent consistency model version of Marigold (Marigold-LCM) to reduce the number of sampling steps significantly. We observe that a single sampling step suffices for our case,
which significantly speeds up inference-time optimization \cite{Novack2024Ditto2, eyring2024reno} relative to the normal 20-50 inference steps \cite{novack2024ditto, Wallace2023EndtoEndDL} that Marigold~\cite{viola2024marigold} uses. We show an ablation study with more sampling steps in appendix.E.
% In contrast, the original Marigold model requires 25-50 inference steps \cite{viola2024marigold} for inference time optimization. 
We also implement custom CUDA kernels for the Disc-PSF forward model. This provides a 2.5x speed up over the PyTorch implementation provided by ~\cite{wang2023implicit} while being more memory efficient, allowing our method to scale to higher-resolution images. Using a single sampling step of Marigold-LCM allows us to compute gradients w.r.t $\zT$ without gradient checkpointing as previously done in \cite{novack2024ditto}. We run the optimization for 200 iterations, which takes roughly 3.5-4 minutes on an NVIDIA A-40 GPU with peak memory usage of 15 GB during the optimization. Please see appendix.A for other implementation details.
\begin{figure}
    \centering
    \includegraphics[width=
    \linewidth]{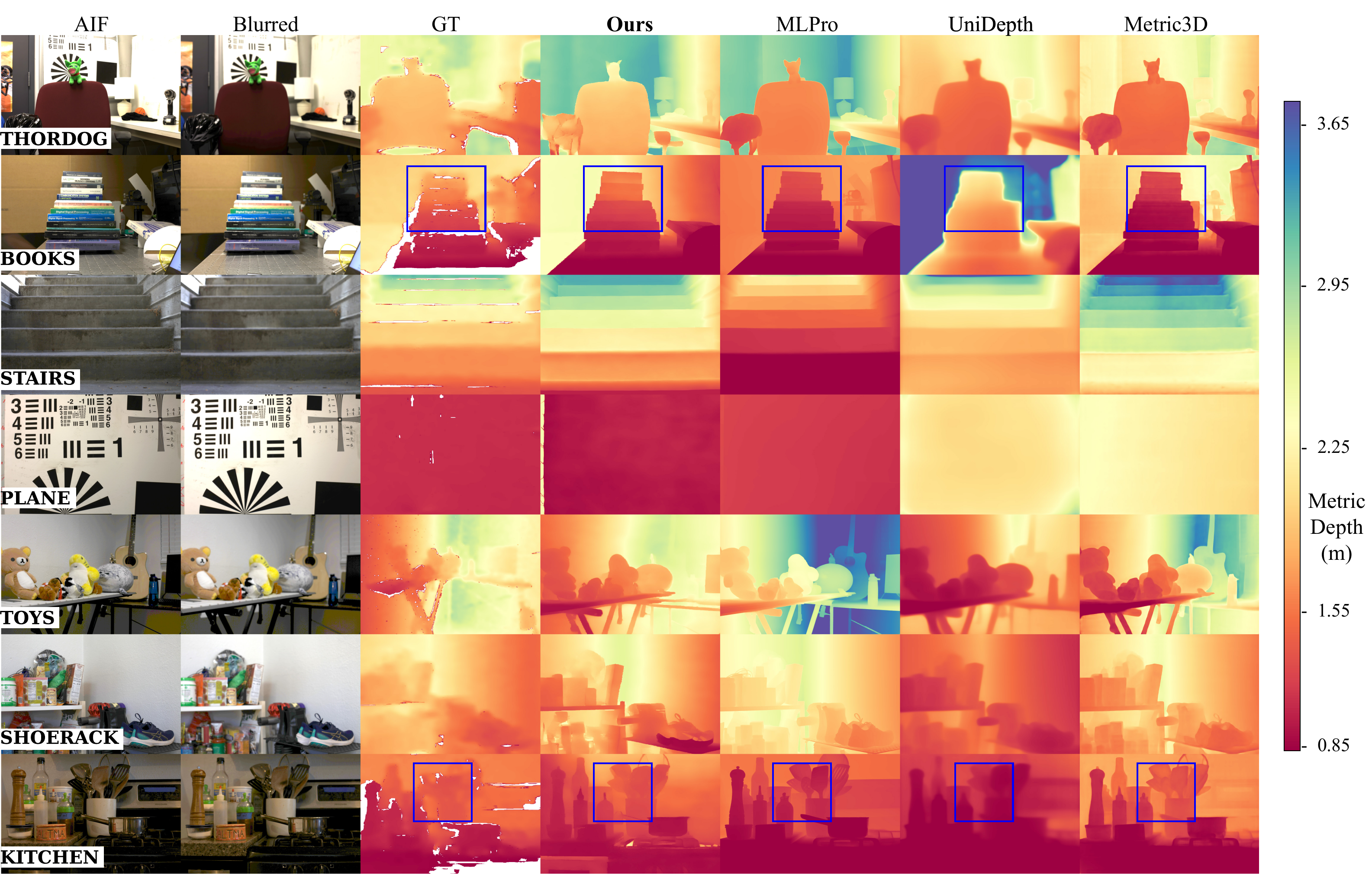} %{figures/real_depth_result.pdf}
\caption{\textbf{Comparisons on our collected dataset.} Our method consistently estimates accurate metric depth across all the scenes. We also observe better relative depth recovery due to leveraging defocus cues (zoom in 4x on blurred) in some regions (blue boxes, \textsc{stairs}). While the competing methods perform comparably to ours in some cases (MLPro:\textsc{plane}, \textsc{kitchen}, UniDepth:\textsc{stairs}, Metric3D:\textsc{thordog}, \textsc{books}), they struggle with the rest of the scenes due to incorrect relative depth (\textsc{toys}, \textsc{books}) and metric scale (\textsc{plane}) recovery. recovers sharp details but fails at metric scale and relative depth accuracy for many of the scenes. Since the RealSense has a wider FOV than the DSLR, we show a roughly aligned crop of the GT depth for comparison.}
    \label{fig:realresults}
\end{figure}
% \paragraph{Implementation Details}
% We use the Adam Optimizer \cite{kingma2014adam} in PyTorch with default hyperparameters. For the real datasets, we use a learning rate of $1.5\times 10^{-3}$ for $\zT$ and $5.0\times10^{-3}$ for $a,b$ respectively.
%We will release the code and dataset in future.
% \begin{itemize}
%     \item Number of sampling steps 
%     \item GPU, Optimizer, learning rates for eps and ab 
%     \item 
%     % \item We implement a disc PSF in CUDA, similar to the forward model in \cite{wang2023implicit}, as the PSF of the camera resembles a disc more than a Gaussian. 
    
% \end{itemize}
\section{Experiments and Results}
% \ct{Nick -- please review the results section and figures, please suggest if we should highlight a certain thing or something.}
\label{sec:experiments}
% \subsection{Results on Simulated data}

% \begin{figure}
%     \centering
%     \includegraphics[width=1.0\linewidth]{figures/nyu_baseline_comparison.png}
%     \caption{NYU Dataset Comparison}
%     \label{fig:enter-label}
% \end{figure}

% \begin{figure}
%     \centering
%     \includegraphics[width=1.0\linewidth]{figures/middlebury_baseline_comparison.png}
%     \caption{Middlebury Dataset Comparison}
%     \label{fig:enter-label}
% \end{figure}

% \subsection{Results on self-collected real dataset}

\begin{figure}
    \centering
    \includegraphics[width=0.4\linewidth]{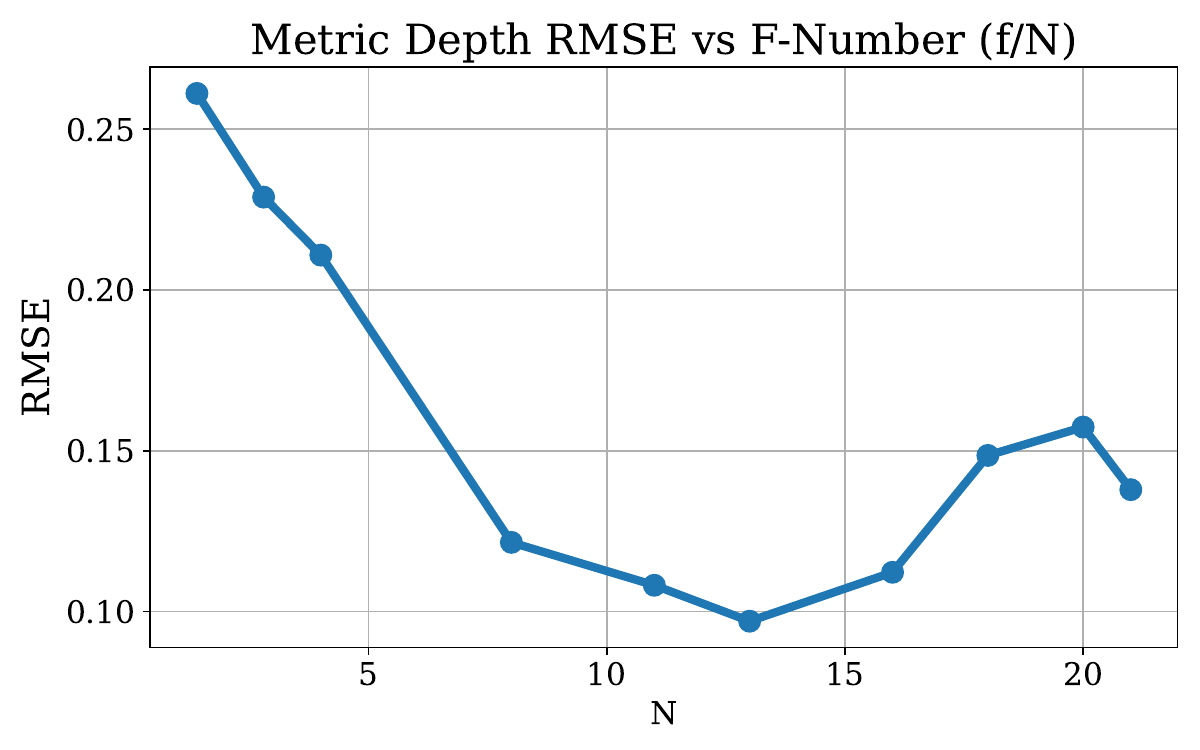}
    \includegraphics[width=0.4\linewidth]{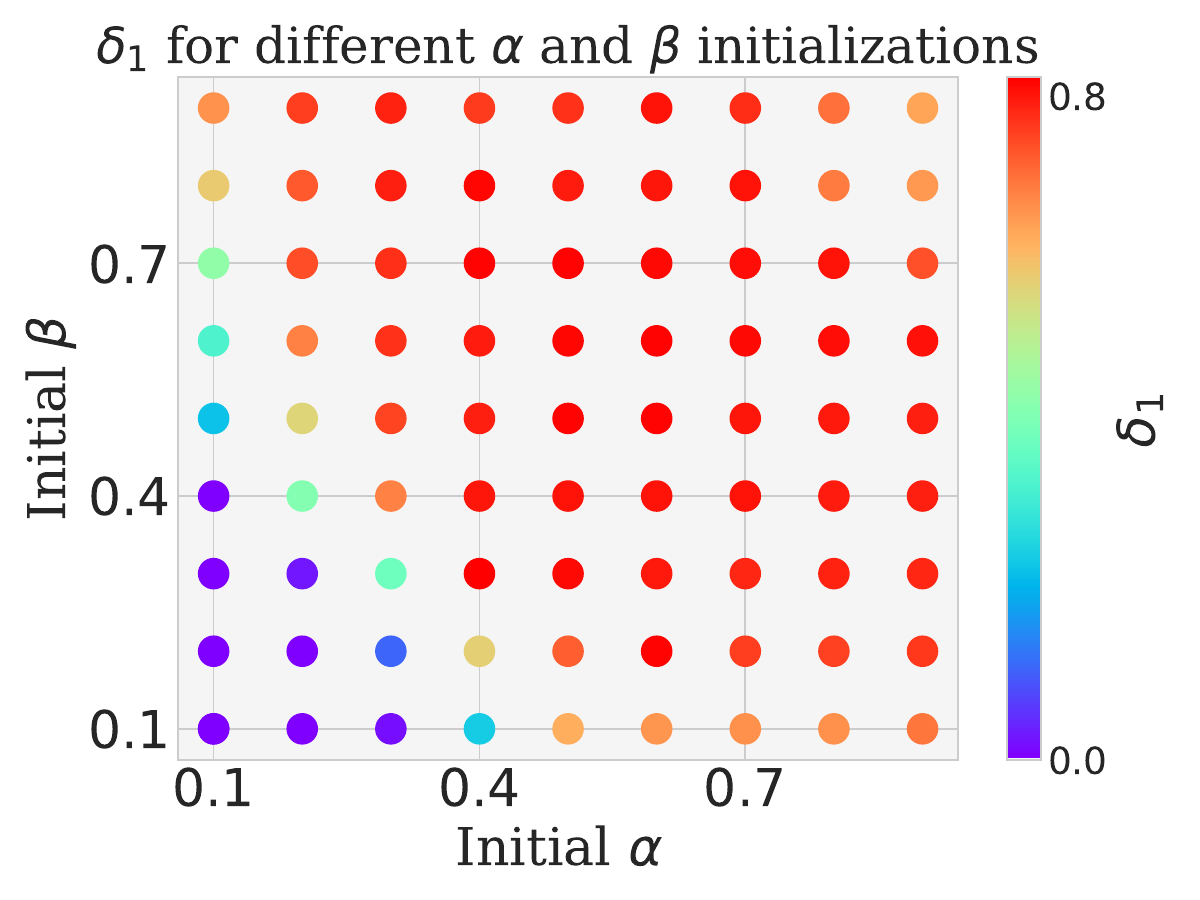}
    \caption{\textbf{Analyzing the effect of different aperture sizes and initializations.} \textit{Left:} We use our forward model to simulate blurred images of a scene from the NYU-v2 dataset. We observe minimum depth error at $N=13$, with errors increasing at more extreme aperture values. \textit{Right:} We plot $\delta_1$ at the end of optimization for various $\alpha,\beta$ initialization (normalized to 0-1). While the performance degrades for small values of $\alpha, \beta$ (bottom left), it is relatively stable for a broad range of initializations. }
    \label{fig:diff-aperture-synth}
\end{figure}

% \na{It looks to me like taking MLPro's results and rendering them through the camera model, then backpropping just to solve for scale and offset would perform pretty well on a lot of our test scenes. Do we have a solid real-world example where MLPro's relative depth makes a major mistake? It makes a pretty good guess on the resolution chart, which is our only "textured plane" example. If we want to avoid that question, we really need to point out a failure case of MLPro where our succeeds} 
\vspace{-1mm}
\paragraph{Dataset details}  Our method requires 2 images captured with different apertures (but same viewpoint) to integrate defocus cues. Standard monocular depth datasets \cite{Scharstein:CVPR:2003, Silberman:ECCV12} typically capture in-focus images at a single aperture per scene, making them unsuitable for evaluating our method. While one could simulate defocused images with \cref{eq:forward_model} and RGBD data, this does not capture the model mismatch (occlusion, diffraction) in the physical image formation process. Therefore, to fairly evaluate our method, we construct a hardware capture setup, shown in  Fig. \cref{fig:psfcomparison}, comprising an Intel RealSense depth camera rigidly mounted to a DSLR camera (Canon EOS 5D Mark II). We use this system to collect a custom real-world dataset of 7 unique scenes, evaluating our method against learning-based MMDE baselines. We choose scenes with diverse subjects and depth profiles, placed within the operating depth range of RealSense (0.3–3.8m) to ensure accurate ground truth depth. Note that the CoC changes negligibly with depth beyond these distances, making defocus cues unreliable. For each scene, we capture images at 6 different apertures, $f/4, f/8, f/11, f/13, f/16$, and $f/22$, with the latter serving as the AIF image $\I$. The blurred image, $\Ib$, is selected from the lower F-stop images (see Fig. \ref{fig:diff-aperture-synth} for comparison of F-stop setting on depth map quality). As described in \cref{sec:method}, we try to maintain a similar ratio of the exposure time and the camera aperture area for all the measurements taken for a scene. The lens focal length ($f$) and F-stop ($N$) are provided by the camera EXIF data, and the focus distance $F$ is read manually from the lens's analog focus scale\footnote{Alternatively, a lens with focus motor encoding would allow this to be known from EXIF}. Please see appendix.B for camera parameters and other dataset details. %\na{Cite specific section}
\paragraph{Evaluation Metrics} We evaluate the predicted metric depth, $\hat{\mathbf{d}}^{\text{m}}$, from our method against the RealSense ground truth depth $\mathbf{d}$ (with overloaded notation), using the metrics in~\cite{bhat2023zoedepth}. Specifically, we compute absolute relative error (REL) $=\frac{1}{M} \sum_{i=1}^{M}\frac{|\mathbf{d}_i - \hat{\mathbf{d}}_i^\text{m}|}{\mathbf{d}_i}$, root mean squared error (RMSE) $= \sqrt{\frac{1}{M} \sum_{i=1}^{M} |\mathbf{d}_{i} - \mathbf{\hat{d}}_{i}^{\text{m}}|^2}$, average log error (log10) $=\frac{1}{M} \sum_{i=1}^{M} |\log_{10} \mathbf{d}_{i} - \log_{10}\mathbf{\hat{d}}_{i}^{\text{m}}|$, and accuracy thresholds $\delta_n =$ fraction of pixels where $\max \left(\frac{\mathbf{d}_i}{\mathbf{\hat{d}}_i^{\text{m}}}, \frac{\mathbf{\hat{d}}_i^{\text{m}}}{\mathbf{d}_i}\right) < (1.25)^{n}$ for $n=1,2,3$. $M$ denotes the number of pixels, and $\mathbf{d}_i, \mathbf{\hat{d}}_i^{\text{m}}$ denote the RealSense and predicted depths at pixel $i$, respectively. We align the ground truth and predicted depth map, similar to \cite{ghanekar2024passive}, and compute the metrics for pixels with non-zero values across the aligned depth maps. See appendix.C for more details on the depth map alignment/calibration procedure. 
% First, we project the metric depth map $\hat{\mathbf{d}}$ into 3D space, and then re-project the resulting point cloud to the realsense camera coordinate system. We then compute the above metrics for pixels with non-zero depths across both the aligned depth maps. 
% \vspace{-5mm}
\paragraph{Quantitative and qualitative results} 
We evaluate our method against some of the recently popular MMDE methods UniDepth~\cite{piccinelli2024unidepth}, Metric3D~\cite{ yin2023metric3d}, and MLPro~\cite{bochkovskii2024depth} on our collected dataset. Both MLPro and UniDepth are provided with the required camera parameters as input. We outperform existing methods qualitatively (\cref{fig:realresults}) and quantitatively (\cref{tab:depth-comparison}) on all the evaluation metrics, averaged over all 7 scenes in the collected dataset. Please see appendix.D for per-scene quantitative metrics.  
% \na{Just overlay on results in fig 4?}
The MMDE methods (MLPro, Metric3D) recover sharper details and are on-par with our method on some scenes (see \cref{fig:realresults}), but they lack consistency in their overall performance across all scenes. Our method achieves better consistency across varying scene conditions. Leveraging defocus cues enables our approach to recover the correct depth scale while also resolving relative depth errors in some cases (\cref{fig:realresults} insets).
% (\ct{show some example of this}). 
We also evaluate our method (row 4 in \cref{tab:depth-comparison}) using a Gaussian PSF \cite{gur2019single} in the forward model. While previous work \cite{gur2019single} uses the Gaussian PSF for training models for unsupervised depth recovery, we find that the model mismatch between the Gaussian PSF and the PSF of the real camera (\cref{fig:psfcomparison}) leads to severe performance degradation in our case compared to using the Disc PSF, which matches the real camera PSF better. This highlights the value of a physically consistent forward model, even with strong learned priors.
% \na{Two ideas for some last-minute stretch results: 1. Could you use your depth map to synthesize blur at a larger aperture than what you captured, then compare that to the real large aperture measurement? We should see that qualitatively and quantitatively we do far better at this than the other methods, and (I think) all you need is a forward pass through your model. 2. Wouldn't your framework apply to different focal lengths as well as different apertures? If it's doable, showing consistent depth images with a few different focal lengths would be nice.}
% \item A uniform disc PSF approximates the camera PSF much better than a Gaussian, specifically for higher blur (see Fig xx). 
% \item On using the DiscPSF as a forward model instead of the Gaussian PSF, we observe significant improvements in the recovered depth.

\begin{table}[h]
\centering
\small
\begin{tabular}{l|cccccc}
\toprule
\textbf{Method} & \textbf{RMSE $\downarrow$} & \textbf{REL $\downarrow$} & \textbf{log10 $\downarrow$} & $\boldsymbol{\delta_1}~\uparrow$ & $\boldsymbol{\delta_2}~\uparrow$ & $\boldsymbol{\delta_3}~\uparrow$ \\
\midrule
MLPro   & 0.468 & 0.246 &	0.105 & 0.597 & 0.821	& 0.990 \\
UniDepth  & 0.644 & 0.376 & 0.157 & 0.259 & 0.684 & 0.954 \\
Metric3D  & 0.459 & 0.295 & 0.106 & 0.650 & 0.825 & 0.895 \\
Ours - Gaussian & 0.528 & 0.279 & 0.142 & 0.422 & 0.695 & 0.928 \\
\textbf{Ours - Disc}     & \textbf{0.273} & \textbf{0.125} & \textbf{0.052} & \textbf{0.879} & \textbf{0.975} & \textbf{0.991} \\
\bottomrule
\end{tabular}
\caption{\textbf{Comparison with learning based MMDE methods.} Our method with the Disc PSF outperforms all the MMDE baselines averaged over all scenes in our dataset. The disc PSF, being more consistent with the real camera PSF, outperforms the Gaussian PSF.(last row).}
% Metrics with $\downarrow$ are better when lower; metrics with $\uparrow$ are better when higher.}
\label{tab:depth-comparison}
\vspace{-4mm}
\end{table}
\subsection{Ablation Studies}
\begin{figure}
    \centering
    \includegraphics[width=\linewidth]{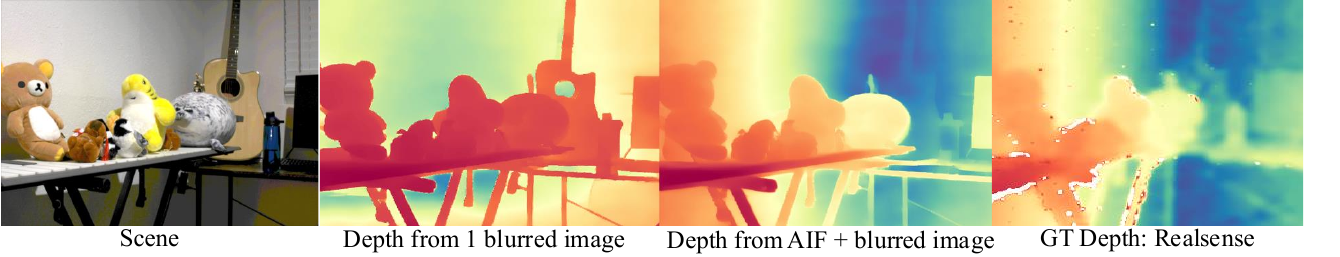}
    \caption{\textbf{Degradation in depth quality on using only a single blurred image.} For the \textsc{toys} scene, we observe that using a single blurred image as input results in severely inaccurate relative depth (middle), with all toys, guitar, and the monitor at similar relative depths. Our proposed method (right) recovers the depth ordering between the objects more accurately.} %\na{Fix crop on GT depth}}
    \label{fig:ablation1inp}
    % \na{Include GT}
    \vspace{-5mm}
\end{figure}

\paragraph{Only single blurred image as input} To test the strength of the diffusion prior, we evaluate our method without using the AIF image. We use a single modestly blurred image ($f/16$) from \textsc{toys} as input to both Marigold and the loss function in (\cref{eq:optimizationobj}), while retaining the previous optimization settings.
% We use the () for this evaluation, with  as our proposed method.
% Given the strength of the diffusion prior, we assess our method with only a single modestly blurred image ($f/16$) as input to both marigold and the optimization objective in (\cref{eq:optimizationobj}, without using the AIF as the second image. 
% which uses both the blurred image and AIF.
This results in a significant degradation in RMSE (1.36), compared to our proposed method (0.346). We visually compare the inaccuracies from this ablation with our method in \cref{fig:ablation1inp}.
% for   While the loss decreases throughout refinement, the depth converges to an incorrect solution, see \cref{fig:ablation1inp}.
\paragraph{Sensitivity to $\alpha,\beta$ initialization:} 
Gradient-based methods are known to be susceptible to local minima for non-convex optimizations. We thus evaluate the sensitivity of our method to the initialization for $\alpha, \beta$. We run the optimization in (\cref{eq:optimizationobj}) with a fixed $\mathbf{z}_{T}^{(\mathbf{d})}$ while grid-searching over initial values of $\alpha$ and $\beta$. In \cref{fig:diff-aperture-synth} (right), we observe that performance (measured by $\delta_1$) drops for small initialization values but remains stable across a broad range around $\alpha=0.5, \beta=0.5$, supporting the robustness of our chosen initialization. We also ablate on sensitivity to initial $\zT$ in appendix.E.
\paragraph{Improvements in relative depth from defocus cues} 
We quantitatively evaluate the effect of defocus cues in our method on improving the relative depth quality. Optimizing only $\alpha,\beta$ while holding $\zT$ constant leads to a performance drop (\cref{tab:ablation-comparison}). This highlights the role of defocus cues in refining the relative depth initially predicted by Marigold. Please see appendix.F for visualizations. 
% We also compare our relative depth quality to MLPro, which produces visually sharp depth boundaries (\cref{fig:realresults}). We normalize the MLPro output between 0 and 1 (normalizing global offsets), and perform a grid search on the affine scaling parameters to find the optimal $\alpha,\beta$ values that best fit the RealSense depth. The results (row 2 in \cref{tab:ablation-comparison}) improve compared to MLPro but are still worse than our method. This shows that visually sharp depth maps do not always imply relative depth accuracy.
\begin{table}[h]
\centering
\small
\begin{tabular}{l|cccccc}
\toprule
\textbf{Method} & \textbf{RMSE $\downarrow$} & \textbf{REL $\downarrow$} & \textbf{log10 $\downarrow$} & $\boldsymbol{\delta_1}~\uparrow$ & $\boldsymbol{\delta_2}~\uparrow$ & $\boldsymbol{\delta_3}~\uparrow$ \\
\midrule
% MLPro   & 0.468 & 0.246 &	0.105 & 0.597 & 0.821	& 0.990 \\ 
% MLPro*     & 0.412 & 0.201 & 0.101 & 0.647 & 0.837 & 0.944 \\ 
Ours $\alpha,\beta$ opt  & 0.297 & 0.156 & 0.069 & 0.743 & 0.957 & 0.99 \\
% \hline 
\textbf{Ours} & \textbf{0.273} & \textbf{0.125} &	\textbf{0.052} & \textbf{0.879} & \textbf{0.975}	& \textbf{0.991}\\
% \hline
% Metric3D  & 0.459 & 0.295 & 0.106 & 0.650 & 0.825 & 0.895 \\
% \hline
% Ours - Gaussian & 0.528 & 0.279 & 0.142 & 0.422 & 0.695 & 0.928 \\
% \hline
% \textbf{Ours - Disc}     & \textbf{0.273} & \textbf{0.125} & \textbf{0.052} & \textbf{0.879} & \textbf{0.975} & \textbf{0.991} \\
\bottomrule
\end{tabular}
\caption{\textbf{Relative depth quality} Optimizing with the noise latent with defocus cues (ours) improves upon optimizing only the affine parameters ($\alpha,\beta$ opt). }
% Metrics with $\downarrow$ are better when lower; metrics with $\uparrow$ are better when higher.}
\label{tab:ablation-comparison}
\end{table}
\vspace{-7mm}

\paragraph{Different aperture sizes} 
We analyze how the aperture (F-stop) used for capturing the blurred image affects our performance. To do this, we use a scene from the NYU-v2 \cite{Silberman:ECCV12}, an indoor RGBD dataset with high-quality ground truth depth annotations. This synthetic setup allows evaluating large F-stops that cannot be captured with our camera, while isolating aperture size from forward model mismatches in a real setup.
Using the ground truth depth and our forward model (\cref{eq:forward_model}), we simulate the blurred images $\Ib$ at varying F-stops ($N$ values) and compute the error metrics between the ground truth and our predicted depth.  We observe in \cref{fig:diff-aperture-synth} that the performance (measured in RMSE) degrades for extreme aperture sizes. This is expected, as extreme blur (high or low) makes the inverse problem ill-posed, and an optimal blur level is key for accurate depth recovery. While $N=13$ appears to be optimal in simulation (ignoring model mismatch), it underperforms $N=8$ on average for real scenes, likely due to low contrast and insufficient blur cues in some of the scenes (\textsc{stairs}, \textsc{plane}). Please see appendix.G for results across all apertures captured in the real dataset.
% compute depth errorfrom our method errors at each of the blur levels errors of We simulate the blurred image ($\Ib$) at varying F-stops using the ground truth depth and our forward model (\cref{eq:forward_model}), 
% This is expected as extreme blur (high or low) makes the inverse problem very ill-posed. An optimal blur level is necessary for the best depth recovery. 
% we simulate different blur levels ( ignoring forward model mismatch) using the Disc-PSF forward model, and use our method to recover the metric depth.
% We solve for metric depth with input blurred images simulated at varying F-stops ($N$ values), to evaluate the effect of the aperture size on the depth accuracy. we simulate different blur levels ( isolating camera non-idealities and forward model mismatch) using the Disc-PSF forward model, and use our method to recover the metric depth. We observe in \cref{fig:diff-aperture-synth} that the performance (measured in RMSE) degrades for very large or small apertures. This is expected as extreme blur (high or low) makes the inverse problem very ill-posed. An optimal blur level is necessary for the best depth recovery. 

% Ran a scene with all the captured apertures as blurred images. Seems like f/4 does the best. The performance worsens as the amount of blur increases. However, we need to do this on synthetic data with really low (F/1.4) to see how well that works and if the trend holds up. This is WIP. 
\section{Limitations and Future Work}
\label{sec:limitations}
While our method outperforms data-driven MMDE baselines, it remains significantly slower at inference time.  Our method is best suited to indoor scenes with a small depth range for which defocus blur offers high depth sensitivity. We observe that if the initial Marigold prediction is severely incorrect in some regions (visualized in appendix.F), the optimization may not always be able to fully correct them (ours for \textsc{shoerack}, \textsc{thordog} in \cref{fig:realresults}). A possible extension of our method is to jointly estimate the AIF and depth map from a single blurred input, as previously explored in approaches~\cite{ghanekar2024passive, abuolaim2020defocus} that are not zero-shot. Our framework can broaden the utility of pre-trained depth priors to scientific applications involving depth-dependent imaging processes such as hyperspectral imaging \cite{kho2019imaging}, endoscopy \cite{liu2023self}, and microscopy \cite{yanny2020miniscope3d}. While we avoid coarsely discretizing the depth map \cite{ikoma2021depth}, our forward model loses accuracy at occlusion boundaries. We envision further improvements through better PSF engineering (coded aperture masks) and more accurate forward modeling of defocus blur.
% \begin{itemize}
%     \item Runtime not instantaneous, compared to MMDE methods 
%     \item can improve further by using an octagonal PSF
%     \item Limited to indoor scenes, but can potentially work with any other differentiable forward model scenario.
%     \item with better PSF engineering, this might lead to even better results.
%     \item While we don't bin the depth map in discrete bins, the forward model would lose accuracy at occlusion boundaries (cite).  
% \end{itemize}
% \section{Conclusion}
% \label{sec:conclusion}

\paragraph{Acknowledgements} We thank Alankar Kotwal for helpful discussions, and Agastya Kalra for proofreading the draft. 
\begin{small}
\bibliographystyle{plain}
\bibliography{sample}

\begin{thebibliography}{10}

\bibitem{abuolaim2020defocus}
Abdullah Abuolaim and Michael~S Brown.
\newblock Defocus deblurring using dual-pixel data.
\newblock In {\em Computer Vision--ECCV 2020: 16th European Conference, Glasgow, UK, August 23--28, 2020, Proceedings, Part X 16}, pages 111--126. Springer, 2020.

\bibitem{alexander2016focal}
Emma Alexander, Qi~Guo, Sanjeev Koppal, Steven Gortler, and Todd Zickler.
\newblock Focal flow: Measuring distance and velocity with defocus and differential motion.
\newblock In {\em Computer Vision--ECCV 2016: 14th European Conference, Amsterdam, The Netherlands, October 11-14, 2016, Proceedings, Part III 14}, pages 667--682. Springer, 2016.

\bibitem{antipa2017diffusercam}
Nick Antipa, Grace Kuo, Reinhard Heckel, Ben Mildenhall, Emrah Bostan, Ren Ng, and Laura Waller.
\newblock Diffusercam: lensless single-exposure 3d imaging.
\newblock {\em Optica}, 5(1):1--9, 2017.

\bibitem{barany2007gaussian}
Imre B{\'a}r{\'a}ny, Van Vu, et~al.
\newblock Central limit theorems for gaussian polytopes.
\newblock {\em The Annals of Probability}, 35(4):1593--1621, 2007.

\bibitem{bhat2023zoedepth}
Shariq~Farooq Bhat, Reiner Birkl, Diana Wofk, Peter Wonka, and Matthias M{\"u}ller.
\newblock Zoedepth: Zero-shot transfer by combining relative and metric depth.
\newblock {\em arXiv preprint arXiv:2302.12288}, 2023.

\bibitem{birkl2023midas}
Reiner Birkl, Diana Wofk, and Matthias M{\"u}ller.
\newblock Midas v3. 1--a model zoo for robust monocular relative depth estimation.
\newblock {\em arXiv preprint arXiv:2307.14460}, 2023.

\bibitem{bochkovskii2024depth}
Aleksei Bochkovskii, Ama{\"e}l Delaunoy, Hugo Germain, Marcel Santos, Yichao Zhou, Stephan~R Richter, and Vladlen Koltun.
\newblock Depth pro: Sharp monocular metric depth in less than a second.
\newblock {\em arXiv preprint arXiv:2410.02073}, 2024.

\bibitem{carvalho2018deep}
Marcela Carvalho, Bertrand Le~Saux, Pauline Trouv{\'e}-Peloux, Andr{\'e}s Almansa, and Fr{\'e}d{\'e}ric Champagnat.
\newblock Deep depth from defocus: how can defocus blur improve 3d estimation using dense neural networks?
\newblock In {\em Proceedings of the European Conference on Computer Vision (ECCV) Workshops}, pages 0--0, 2018.

\bibitem{chung2023diffusion}
Hyungjin Chung, Jeongsol Kim, Michael~T. McCann, Marc~L. Klasky, and Jong~Chul Ye.
\newblock Diffusion posterior sampling for general noisy inverse problems.
\newblock In {\em International Conference on Learning Representations (ICLR)}, 2023.

\bibitem{chung2022improving}
Hyungjin Chung, Byeongsu Sim, Dohoon Ryu, and Jong~Chul Ye.
\newblock Improving diffusion models for inverse problems using manifold constraints.
\newblock In {\em Advances in Neural Information Processing Systems (NeurIPS)}, 2022.

\bibitem{chung2022come}
Hyungjin Chung, Byeongsu Sim, and Jong~Chul Ye.
\newblock Come-closer-diffuse-faster: Accelerating conditional diffusion models for inverse problems through stochastic contraction.
\newblock In {\em Proceedings of the IEEE/CVF Conference on Computer Vision and Pattern Recognition (CVPR)}, pages 12403--12412, 2022.

\bibitem{daras2024survey}
Giannis Daras, Hyungjin Chung, Chieh-Hsin Lai, Yuki Mitsufuji, Jong~Chul Ye, Peyman Milanfar, Alexandros~G Dimakis, and Mauricio Delbracio.
\newblock A survey on diffusion models for inverse problems.
\newblock {\em arXiv preprint arXiv:2410.00083}, 2024.

\bibitem{duan2024diffusiondepth}
Yiquan Duan, Xianda Guo, and Zheng Zhu.
\newblock Diffusiondepth: Diffusion denoising approach for monocular depth estimation.
\newblock In {\em European Conference on Computer Vision}, pages 432--449. Springer, 2024.

\bibitem{eyring2024reno}
Luca Eyring, Shyamgopal Karthik, Karsten Roth, Alexey Dosovitskiy, and Zeynep Akata.
\newblock Reno: Enhancing one-step text-to-image models through reward-based noise optimization.
\newblock {\em Advances in Neural Information Processing Systems}, 37:125487--125519, 2024.

\bibitem{farid1998range}
Hany Farid and Eero~P Simoncelli.
\newblock Range estimation by optical differentiation.
\newblock {\em Journal of the Optical Society of America A}, 15(7):1777--1786, 1998.

\bibitem{fu2024geowizard}
Xiao Fu, Wei Yin, Mu~Hu, Kaixuan Wang, Yuexin Ma, Ping Tan, Shaojie Shen, Dahua Lin, and Xiaoxiao Long.
\newblock Geowizard: Unleashing the diffusion priors for 3d geometry estimation from a single image.
\newblock In {\em European Conference on Computer Vision}, pages 241--258. Springer, 2024.

\bibitem{Gal2022RareConcepts}
Rinon Gal, Yuval Alaluf, Yuval Atzmon, Or~Patashnik, Amit~H. Bermano, Tal Hassner, and Daniel Cohen-Or.
\newblock An image is worth one word: Personalizing text-to-image generation using textual inversion.
\newblock {\em arXiv preprint arXiv:2208.01618}, 2022.

\bibitem{garg2019learning}
Rahul Garg, Neal Wadhwa, Sameer Ansari, and Jonathan~T Barron.
\newblock Learning single camera depth estimation using dual-pixels.
\newblock In {\em Proceedings of the IEEE/CVF international conference on computer vision}, pages 7628--7637, 2019.

\bibitem{ghanekar2024passive}
Bhargav Ghanekar, Salman~Siddique Khan, Pranav Sharma, Shreyas Singh, Vivek Boominathan, Kaushik Mitra, and Ashok Veeraraghavan.
\newblock Passive snapshot coded aperture dual-pixel rgb-d imaging.
\newblock In {\em Proceedings of the IEEE/CVF Conference on Computer Vision and Pattern Recognition}, pages 25348--25357, 2024.

\bibitem{gur2019single}
Shir Gur and Lior Wolf.
\newblock Single image depth estimation trained via depth from defocus cues.
\newblock In {\em Proceedings of the IEEE/CVF conference on computer vision and pattern recognition}, pages 7683--7692, 2019.

\bibitem{hazirbas2019deep}
Caner Hazirbas, Sebastian~Georg Soyer, Maximilian~Christian Staab, Laura Leal-Taix{\'e}, and Daniel Cremers.
\newblock Deep depth from focus.
\newblock In {\em Computer Vision--ACCV 2018: 14th Asian Conference on Computer Vision, Perth, Australia, December 2--6, 2018, Revised Selected Papers, Part III 14}, pages 525--541. Springer, 2019.

\bibitem{he2024lotus}
Jing He, Haodong Li, Wei Yin, Yixun Liang, Leheng Li, Kaiqiang Zhou, Hongbo Zhang, Bingbing Liu, and Ying-Cong Chen.
\newblock Lotus: Diffusion-based visual foundation model for high-quality dense prediction.
\newblock {\em arXiv preprint arXiv:2409.18124}, 2024.

\bibitem{ikoma2021depth}
Hayato Ikoma, Cindy~M Nguyen, Christopher~A Metzler, Yifan Peng, and Gordon Wetzstein.
\newblock Depth from defocus with learned optics for imaging and occlusion-aware depth estimation.
\newblock In {\em 2021 IEEE International Conference on Computational Photography (ICCP)}, pages 1--12. IEEE, 2021.

\bibitem{jiang2024construct}
Kaiwen Jiang, Yang Fu, Mukund Varma~T, Yash Belhe, Xiaolong Wang, Hao Su, and Ravi Ramamoorthi.
\newblock A construct-optimize approach to sparse view synthesis without camera pose.
\newblock In {\em ACM SIGGRAPH 2024 Conference Papers}, pages 1--11, 2024.

\bibitem{jones1926relation}
Loyd~A Jones, Emery Huse, and Vincent~C Hall.
\newblock On the relation between time and intensity in photographic exposure.
\newblock {\em Journal of the Optical Society of America}, 12(4):321--348, 1926.

\bibitem{ke2024repurposing}
Bingxin Ke, Anton Obukhov, Shengyu Huang, Nando Metzger, Rodrigo~Caye Daudt, and Konrad Schindler.
\newblock Repurposing diffusion-based image generators for monocular depth estimation.
\newblock In {\em Proceedings of the IEEE/CVF Conference on Computer Vision and Pattern Recognition}, pages 9492--9502, 2024.

\bibitem{kho2019imaging}
Esther Kho, Lisanne~L de~Boer, Anouk~L Post, Koen~K Van~de Vijver, Katarzyna J{\'o}{\'z}wiak, Henricus~JCM Sterenborg, and Theo~JM Ruers.
\newblock Imaging depth variations in hyperspectral imaging: development of a method to detect tumor up to the required tumor-free margin width.
\newblock {\em Journal of biophotonics}, 12(11):e201900086, 2019.

\bibitem{levin2007image}
Anat Levin, Rob Fergus, Fr{\'e}do Durand, and William~T Freeman.
\newblock Image and depth from a conventional camera with a coded aperture.
\newblock {\em ACM transactions on graphics (TOG)}, 26(3):70--es, 2007.

\bibitem{lin2015depth}
Haiting Lin, Can Chen, Sing~Bing Kang, and Jingyi Yu.
\newblock Depth recovery from light field using focal stack symmetry.
\newblock In {\em Proceedings of the IEEE International Conference on Computer Vision}, pages 3451--3459, 2015.

\bibitem{liu2017matting}
Chao Liu, Srinivasa~G Narasimhan, and Artur~W Dubrawski.
\newblock Matting and depth recovery of thin structures using a focal stack.
\newblock In {\em Proceedings of the IEEE Conference on Computer Vision and Pattern Recognition}, pages 6970--6978, 2017.

\bibitem{liu2023self}
Yuying Liu and Siyang Zuo.
\newblock Self-supervised monocular depth estimation for gastrointestinal endoscopy.
\newblock {\em Computer Methods and Programs in Biomedicine}, 238:107619, 2023.

\bibitem{liu2021swin}
Ze~Liu, Yutong Lin, Yue Cao, Han Hu, Yixuan Wei, Zheng Zhang, Stephen Lin, and Baining Guo.
\newblock Swin transformer: Hierarchical vision transformer using shifted windows.
\newblock In {\em Proceedings of the IEEE/CVF international conference on computer vision}, pages 10012--10022, 2021.

\bibitem{maximov2020focus}
Maxim Maximov, Kevin Galim, and Laura Leal-Taix{\'e}.
\newblock Focus on defocus: bridging the synthetic to real domain gap for depth estimation.
\newblock In {\em Proceedings of the IEEE/CVF conference on computer vision and pattern recognition}, pages 1071--1080, 2020.

\bibitem{nathan2024osmosis}
Opher~Bar Nathan, Deborah Levy, Tali Treibitz, and Dan Rosenbaum.
\newblock Osmosis: Rgbd diffusion prior for underwater image restoration.
\newblock In {\em European Conference on Computer Vision}, pages 302--319. Springer, 2024.

\bibitem{Silberman:ECCV12}
Pushmeet~Kohli Nathan~Silberman, Derek~Hoiem and Rob Fergus.
\newblock Indoor segmentation and support inference from rgbd images.
\newblock In {\em ECCV}, 2012.

\bibitem{Novack2024Ditto2}
Zachary Novack, Julian McAuley, Taylor Berg-Kirkpatrick, and Nicholas~J. Bryan.
\newblock {DITTO-2}: Distilled diffusion inference-time t-optimization for music generation.
\newblock In {\em International Society of Music Information Retrieval (ISMIR)}, 2024.

\bibitem{novack2024ditto}
Zachary Novack, Julian McAuley, Taylor Berg-Kirkpatrick, and Nicholas~J. Bryan.
\newblock {DITTO}: Diffusion inference-time t-optimization for music generation.
\newblock In {\em International Conference on Machine Learning (ICML)}, 2024.

\bibitem{pan2021dual}
Liyuan Pan, Shah Chowdhury, Richard Hartley, Miaomiao Liu, Hongguang Zhang, and Hongdong Li.
\newblock Dual pixel exploration: Simultaneous depth estimation and image restoration.
\newblock In {\em Proceedings of the IEEE/CVF Conference on Computer Vision and Pattern Recognition}, pages 4340--4349, 2021.

\bibitem{piccinelli2024unidepth}
Luigi Piccinelli, Yung-Hsu Yang, Christos Sakaridis, Mattia Segu, Siyuan Li, Luc Van~Gool, and Fisher Yu.
\newblock Unidepth: Universal monocular metric depth estimation.
\newblock In {\em Proceedings of the IEEE/CVF Conference on Computer Vision and Pattern Recognition}, pages 10106--10116, 2024.

\bibitem{10.1145/800224.806818}
Michael Potmesil and Indranil Chakravarty.
\newblock A lens and aperture camera model for synthetic image generation.
\newblock In {\em Proceedings of the 8th Annual Conference on Computer Graphics and Interactive Techniques}, SIGGRAPH '81, page 297–305, New York, NY, USA, 1981. Association for Computing Machinery.

\bibitem{ranftl2020towards}
Ren{\'e} Ranftl, Katrin Lasinger, David Hafner, Konrad Schindler, and Vladlen Koltun.
\newblock Towards robust monocular depth estimation: Mixing datasets for zero-shot cross-dataset transfer.
\newblock {\em IEEE transactions on pattern analysis and machine intelligence}, 44(3):1623--1637, 2020.

\bibitem{rombach2022high}
Robin Rombach, Andreas Blattmann, Dominik Lorenz, Patrick Esser, and Bj{\"o}rn Ommer.
\newblock High-resolution image synthesis with latent diffusion models.
\newblock In {\em Proceedings of the IEEE/CVF conference on computer vision and pattern recognition}, pages 10684--10695, 2022.

\bibitem{rout2024solving}
Litu Rout, Negin Raoof, Giannis Daras, Constantine Caramanis, Alex Dimakis, and Sanjay Shakkottai.
\newblock Solving linear inverse problems provably via posterior sampling with latent diffusion models.
\newblock {\em Advances in Neural Information Processing Systems}, 36, 2024.

\bibitem{samuel2024norm}
Dvir Samuel, Rami Ben-Ari, Nir Darshan, Haggai Maron, and Gal Chechik.
\newblock Norm-guided latent space exploration for text-to-image generation.
\newblock {\em Advances in Neural Information Processing Systems}, 36, 2024.

\bibitem{saxena2005learning}
Ashutosh Saxena, Sung Chung, and Andrew Ng.
\newblock Learning depth from single monocular images.
\newblock {\em Advances in neural information processing systems}, 18, 2005.

\bibitem{saxena2023diffusion}
Saurabh Saxena, Charles Herrmann, Junhwa Hur, Abhishek Kar, Mohammad Norouzi, Deqing Sun, and David~J Fleet.
\newblock The surprising effectiveness of diffusion models for optical flow and monocular depth estimation.
\newblock In A.~Oh, T.~Naumann, A.~Globerson, K.~Saenko, M.~Hardt, and S.~Levine, editors, {\em Advances in Neural Information Processing Systems}, volume~36, pages 39443--39469. Curran Associates, Inc., 2023.

\bibitem{saxena2023zero}
Saurabh Saxena, Junhwa Hur, Charles Herrmann, Deqing Sun, and David~J Fleet.
\newblock Zero-shot metric depth with a field-of-view conditioned diffusion model.
\newblock {\em arXiv preprint arXiv:2312.13252}, 2023.

\bibitem{Scharstein:CVPR:2003}
Daniel Scharstein and Richard Szeliski.
\newblock High-accuracy stereo depth maps using structured light.
\newblock In {\em Proceedings of the IEEE Computer Society Conference on Computer Vision and Pattern Recognition (CVPR)}, volume~1, pages 195--202. IEEE, 2003.

\bibitem{schon2021mgnet}
Markus Sch{\"o}n, Michael Buchholz, and Klaus Dietmayer.
\newblock Mgnet: Monocular geometric scene understanding for autonomous driving.
\newblock In {\em Proceedings of the IEEE/CVF International Conference on Computer Vision}, pages 15804--15815, 2021.

\bibitem{sheng2024dr}
Yichen Sheng, Zixun Yu, Lu~Ling, Zhiwen Cao, Xuaner Zhang, Xin Lu, Ke~Xian, Haiting Lin, and Bedrich Benes.
\newblock Dr. bokeh: Differentiable occlusion-aware bokeh rendering.
\newblock In {\em Proceedings of the IEEE/CVF Conference on Computer Vision and Pattern Recognition}, pages 4515--4525, 2024.

\bibitem{song2023solving}
Bowen Song, Soo~Min Kwon, Zecheng Zhang, Xinyu Hu, Qing Qu, and Liyue Shen.
\newblock Solving inverse problems with latent diffusion models via hard data consistency.
\newblock {\em arXiv preprint arXiv:2307.08123}, 2023.

\bibitem{song2023pseudoinverse}
Jiaming Song, Arash Vahdat, Morteza Mardani, and Jan Kautz.
\newblock Pseudoinverse-guided diffusion models for inverse problems.
\newblock In {\em International Conference on Learning Representations}, 2023.

\bibitem{song2025depthmaster}
Ziyang Song, Zerong Wang, Bo~Li, Hao Zhang, Ruijie Zhu, Li~Liu, Peng-Tao Jiang, and Tianzhu Zhang.
\newblock Depthmaster: Taming diffusion models for monocular depth estimation.
\newblock {\em arXiv preprint arXiv:2501.02576}, 2025.

\bibitem{srinivasan2018aperture}
Pratul~P Srinivasan, Rahul Garg, Neal Wadhwa, Ren Ng, and Jonathan~T Barron.
\newblock Aperture supervision for monocular depth estimation.
\newblock In {\em Proceedings of the IEEE Conference on Computer Vision and Pattern Recognition}, pages 6393--6401, 2018.

\bibitem{strecke2017accurate}
Michael Strecke, Anna Alperovich, and Bastian Goldluecke.
\newblock Accurate depth and normal maps from occlusion-aware focal stack symmetry.
\newblock In {\em Proceedings of the IEEE Conference on Computer Vision and Pattern Recognition}, pages 2814--2822, 2017.

\bibitem{subbarao1994depth}
Murali Subbarao and Gopal Surya.
\newblock Depth from defocus: A spatial domain approach.
\newblock {\em International Journal of computer vision}, 13(3):271--294, 1994.

\bibitem{tang2017depth}
Huixuan Tang, Scott Cohen, Brian Price, Stephen Schiller, and Kiriakos~N Kutulakos.
\newblock Depth from defocus in the wild.
\newblock In {\em Proceedings of the IEEE conference on computer vision and pattern recognition}, pages 2740--2748, 2017.

\bibitem{viola2024marigold}
Massimiliano Viola, Kevin Qu, Nando Metzger, Bingxin Ke, Alexander Becker, Konrad Schindler, and Anton Obukhov.
\newblock Marigold-dc: Zero-shot monocular depth completion with guided diffusion.
\newblock {\em arXiv preprint arXiv:2412.13389}, 2024.

\bibitem{Wallace2023EndtoEndDL}
Bram Wallace, Akash Gokul, Stefano Ermon, and Nikhil~Vijay Naik.
\newblock End-to-end diffusion latent optimization improves classifier guidance.
\newblock {\em 2023 IEEE/CVF International Conference on Computer Vision (ICCV)}, pages 7246--7256, 2023.

\bibitem{wang2023implicit}
Chao Wang, Krzysztof Wolski, Xingang Pan, Thomas Leimk{\"u}hler, Bin Chen, Christian Theobalt, Karol Myszkowski, Hans-Peter Seidel, and Ana Serrano.
\newblock An implicit neural representation for the image stack: Depth, all in focus, and high dynamic range.
\newblock Technical report, 2023.

\bibitem{wang2021bridging}
Ning-Hsu Wang, Ren Wang, Yu-Lun Liu, Yu-Hao Huang, Yu-Lin Chang, Chia-Ping Chen, and Kevin Jou.
\newblock Bridging unsupervised and supervised depth from focus via all-in-focus supervision.
\newblock In {\em Proceedings of the IEEE/CVF international conference on computer vision}, pages 12621--12631, 2021.

\bibitem{watanabe1998rational}
Masahiro Watanabe and Shree~K Nayar.
\newblock Rational filters for passive depth from defocus.
\newblock {\em International Journal of Computer Vision}, 27:203--225, 1998.

\bibitem{wen2025foundationstereo}
Bowen Wen, Matthew Trepte, Joseph Aribido, Jan Kautz, Orazio Gallo, and Stan Birchfield.
\newblock Foundationstereo: Zero-shot stereo matching.
\newblock {\em arXiv preprint arXiv:2501.09898}, 2025.

\bibitem{won2022learning}
Changyeon Won and Hae-Gon Jeon.
\newblock Learning depth from focus in the wild.
\newblock In {\em European Conference on Computer Vision}, pages 1--18. Springer, 2022.

\bibitem{wu2019phasecam3d}
Yicheng Wu, Vivek Boominathan, Huaijin Chen, Aswin Sankaranarayanan, and Ashok Veeraraghavan.
\newblock Phasecam3d—learning phase masks for passive single view depth estimation.
\newblock In {\em 2019 IEEE International Conference on Computational Photography (ICCP)}, pages 1--12. IEEE, 2019.

\bibitem{xin2021defocus}
Shumian Xin, Neal Wadhwa, Tianfan Xue, Jonathan~T Barron, Pratul~P Srinivasan, Jiawen Chen, Ioannis Gkioulekas, and Rahul Garg.
\newblock Defocus map estimation and deblurring from a single dual-pixel image.
\newblock In {\em Proceedings of the IEEE/CVF International Conference on Computer Vision}, pages 2228--2238, 2021.

\bibitem{xiong1993depth}
Yalin Xiong and Steven~A Shafer.
\newblock Depth from focusing and defocusing.
\newblock In {\em Proceedings of IEEE Conference on Computer Vision and Pattern Recognition}, pages 68--73. IEEE, 1993.

\bibitem{yang2024depth}
Lihe Yang, Bingyi Kang, Zilong Huang, Xiaogang Xu, Jiashi Feng, and Hengshuang Zhao.
\newblock Depth anything: Unleashing the power of large-scale unlabeled data.
\newblock In {\em Proceedings of the IEEE/CVF Conference on Computer Vision and Pattern Recognition}, pages 10371--10381, 2024.

\bibitem{yang2024depthv2}
Lihe Yang, Bingyi Kang, Zilong Huang, Zhen Zhao, Xiaogang Xu, Jiashi Feng, and Hengshuang Zhao.
\newblock Depth anything v2.
\newblock {\em arXiv preprint arXiv:2406.09414}, 2024.

\bibitem{yanny2020miniscope3d}
Kyrollos Yanny, Nick Antipa, William Liberti, Sam Dehaeck, Kristina Monakhova, Fanglin~Linda Liu, Konlin Shen, Ren Ng, and Laura Waller.
\newblock Miniscope3d: optimized single-shot miniature 3d fluorescence microscopy.
\newblock {\em Light: Science \& Applications}, 9(1):171, 2020.

\bibitem{yeo2023rapid}
Teresa Yeo, O{\u{g}}uzhan~Fatih Kar, Zahra Sodagar, and Amir Zamir.
\newblock Rapid network adaptation: Learning to adapt neural networks using test-time feedback.
\newblock In {\em Proceedings of the IEEE/CVF International Conference on Computer Vision}, pages 4674--4687, 2023.

\bibitem{yin2023metric3d}
Wei Yin, Chi Zhang, Hao Chen, Zhipeng Cai, Gang Yu, Kaixuan Wang, Xiaozhi Chen, and Chunhua Shen.
\newblock Metric3d: Towards zero-shot metric 3d prediction from a single image.
\newblock In {\em Proceedings of the IEEE/CVF International Conference on Computer Vision}, pages 9043--9053, 2023.

\bibitem{zhang2025surveymonocularmetricdepth}
Jiuling Zhang.
\newblock Survey on monocular metric depth estimation, 2025.

\bibitem{zheng2020joint}
Yucheng Zheng and M~Salman Asif.
\newblock Joint image and depth estimation with mask-based lensless cameras.
\newblock {\em IEEE Transactions on Computational Imaging}, 6:1167--1178, 2020.

\bibitem{Zheng2023TiNOEdit}
Yuxuan Zheng, Yifan Li, Yizhuo Zhang, Yiran Zhang, Lin Zhang, and Lei Zhang.
\newblock {TiNO-Edit}: Timestep and noise optimization for robust diffusion-based image editing.
\newblock {\em arXiv preprint arXiv:2304.06720}, 2023.

\bibitem{zhou2009coded}
Changyin Zhou, Stephen Lin, and Shree Nayar.
\newblock Coded aperture pairs for depth from defocus.
\newblock In {\em 2009 IEEE 12th international conference on computer vision}, pages 325--332. IEEE, 2009.

\end{thebibliography}
\end{small}
\newpage
\appendix

\section*{Appendix}
\addcontentsline{toc}{section}{Supplementary Material}
We organize the appendix as follows. In \cref{sec:sec1} (\orange{L214}, main text\footnote{reference line numbers in main text will be denoted in orange color}), we provide implementation details for our method. We outline the dataset capture parameters in \cref{sec:sec2} (\orange{L233}), the calibration procedure to align RealSense and predicted depth in \cref{sec:sec3}(\orange{L241}), and the per-scene quantitative metrics for the scenes in Fig.4 of the main paper in \cref{sec:sec4} (\orange{L247}). We report some additional ablation studies in \cref{sec:sec5} (\orange{L208, L269}). In \cref{sec:sec6}, we visualize how incorporating defocus cues through our optimization affects the relative depth predicted by Marigold (\orange{L273, L291}). In \cref{sec:sec7}, we demonstrate the results from using for blurry images for all the F-stops (besides F/8 used in the paper)  captured as part of the real dataset.

\section{Implementation details}
\label{sec:sec1}
 We use RGB images of size $ 750\times1126$, and depth maps captured at $480\times640$ resolution for evaluation in our method.  We use the Adam optimizer with a learning rate of $ 1.5\times10^ {-3}$ for $\zT$, and $5\times10^{-3}$ for $a,b$, and default values for optimizer parameters. In the CUDA implementation of the Disc PSF,  we use a maximum window size (extent of $i,j$ in equation 3 of the main paper) of 63 pixels. The scene bounds are set to $s_{\min}=1.49$, $s_{\max}=3.5$ during optimization in our method. These values represent a conservative upper bound on the potential maximum scale and offset in the real dataset. Note that we use the \textit{same} scene bounds during optimization for all the real scenes.

% Add all hyperparameter details, image sizes, while optimizing, and downsampling. 

\section{Real Dataset details}
\label{sec:sec2}
\paragraph{RGB Camera details }We used a Canon EOS 5D Mark II as the RGB camera. This camera features a 21MP sensor with dimensions of 5616 x 3744 pixels and a pixel pitch of 6.41 $\mu$m. The sensor uses an RGGB Bayer pattern. We used a Canon lens with a focal length ($f$ in main text) of 50mm. This lens allows adjusting the F-stop from $N=1.4$ to $N=22$. We set the focus distance ($F$) to 80cm for all the captures. 

\paragraph{Depth Camera details} We used an Intel RealSense D435 to capture groundtruth depth. This camera has a stereoscopic depth sensor (with FOV 91.2 degrees) and a 2MP RGB camera. The depth sensor has less than 2\% error at 2 meters (i.e, up to 4cm error at 2 meters). To reduce flying pixel errors in the captured depth maps (due to low texture/illumination, multipath interference, etc), we average the depth maps over 60 frames. 
%     \item To remove noise from the captured depth images, we capture 60 frames and average the depth map to remove noise.
% \item All scenes, where they were captured, peculiarities. 
% \item Add a table with all capture parameters. 
% \item 
We have uploaded the dataset at this \href{https://drive.google.com/drive/folders/1EKlT9LLfHVa8dpB2oEFkOUCXVrEuYA5P?usp=sharing}{link}. For the \textsc{thordog} scene, F/11 and F/16 images are unavailable (due to image corruption), but all the other F-stops are present in the dataset. Please see the README.md file in the dataset for further details on usage. 
\section{Calibration process to align RealSense and Predicted Depth}
\label{sec:sec3}
\begin{figure}
    \centering
    \includegraphics[width=0.5\linewidth]{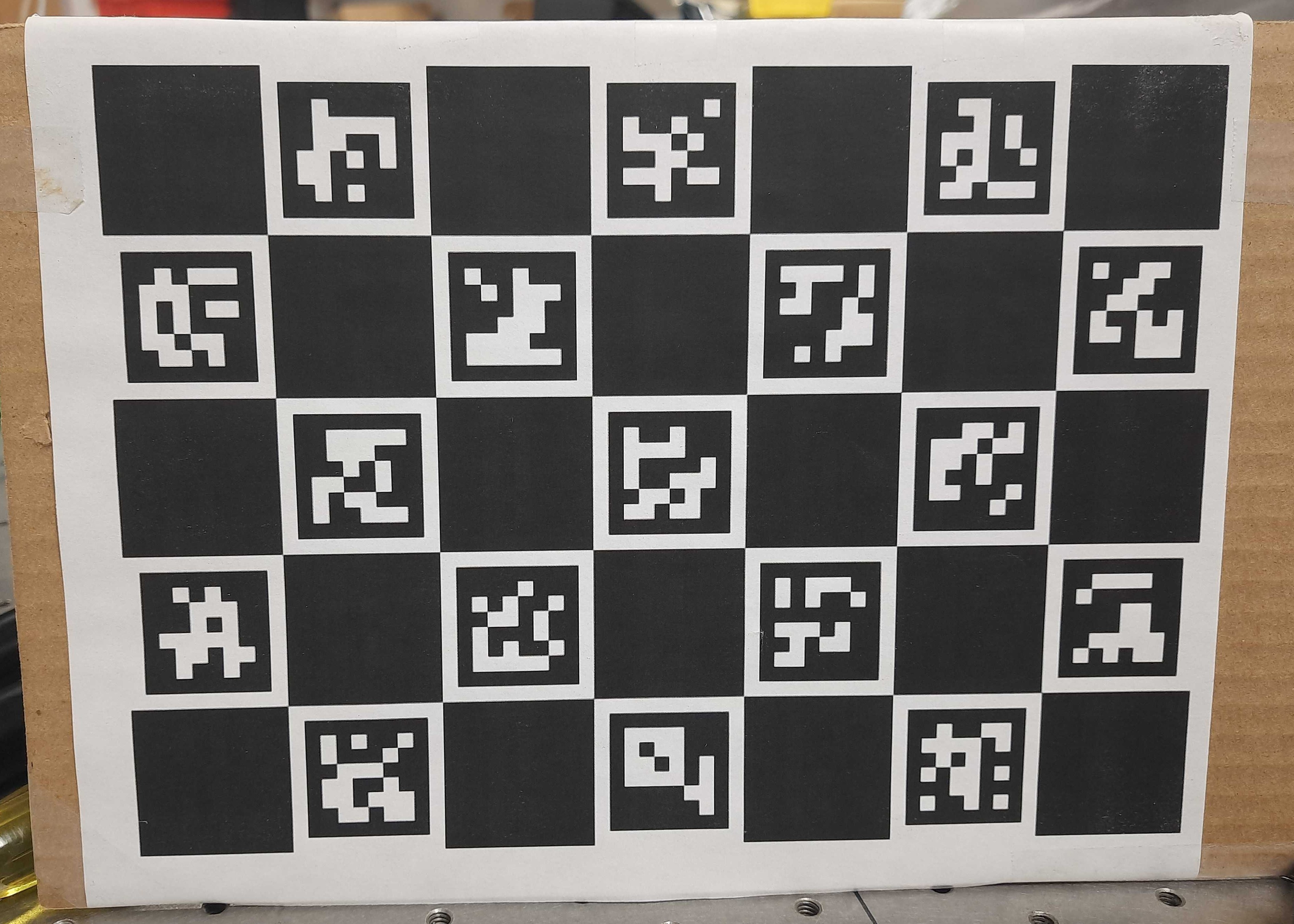}
    \caption{7x5 ChArUco Board}
    \label{fig:charuco}
\end{figure}

Our capture setup consists of the RealSense mounted rigidly to the flash mount of the DSLR (Fig. 2c, main text). Hence, for using the ground truth depth map from RealSense for evaluation, we estimate the pose of the DSLR with respect to the RealSense using camera calibration. We capture 10-15 images of a charuco board (\cref{fig:charuco}), sufficiently covering the depth and XY range. using the DSLR and the RGB camera of RealSense. Note that we use full-resolution images from the DSLR for calibration, and then later rescale the camera intrinsics appropriately. Using OpenCV functions, we estimate the relative pose transformation between the DSLR and the RealSense Camera. We use the \textit{pyrealsense2} library for transforming the captured depth map from the RealSense Depth camera to the RealSense RGB camera pose. We will release the calibration code post acceptance. The above procedure provides us the pose transformation from DSLR to RealSense, denoted as $\mathcal{T}^{\text{RS}}_{\text{DSLR}} = \begin{bmatrix}
    R^{RS}_{DSLR} & T^{RS}_{DSLR}\\ \mathbf{0} & 1
\end{bmatrix} \in \mathbb{R}^{4\times4}$, where $R^{RS}_{DSLR}, T^{RS}_{DSLR}$ correspond to the $3\times3$ rotation matrix and $3\times1$ translation vector.

\paragraph{Aligning depth maps for evaluation}
Using the camera intrinsic matrix $K_{DSLR}$ for the DSLR (estimated during calibration), we transform our predicted depth map (in DSLR coordinate system) to a point cloud $\mathbf{P}_{\text{DSLR}}$ in the DSLR coordinate system. 
\begin{align}
\mathbf{P}_{\text{DSLR}}(x, y, z) = D(u, v) \cdot K_{\text{DSLR}}^{-1} \begin{bmatrix} u \\ v \\ 1 \end{bmatrix},
\end{align}

where $D(u, v)$ is the predicted depth at pixel $(u, v)$. We then transform $\mathcal{P}_{DSLR}$ to the RealSense coordinate system: 
\begin{align}
\mathbf{P}_{\text{RS}}= R^{RS}_{DSLR} \mathbf{P}_{\text{DSLR}} + T^{RS}_{DSLR}.
\end{align}
We then project the transformed point cloud $\mathbf{P}_{\text{RS}}$ to the RealSense Camera plane using the camera projection equation to get the predicted depth map from the perspective of the RealSense camera. Note that since the DSLR has a much shorter FOV than the RealSense camera, the projected depth map (in RealSense coordinate system) will be sparser. We thus compute the metrics only for the pixels at which both ground truth and predicted depth maps have non-zero values. Note that we also undistort the RealSense depth map using the \texttt{cv2.undistort} function in OpenCV before evaluation, where the distortion coefficients are estimated during the camera calibration procedure by the OpenCV API.

% \begin{itemize}
%     \item The depth sensor and the DSLR camera do not have the same pose. Thus, we are required to transform the captured depth image to the DSLR frame of reference.
%     \item There are two levels of transform in our calibration - transform to Intel RealSense's RGB module, and transform from the RGB module to DSLR frame of reference.
%     \item The first transform with in the Intel Realsense, can be performed using the `pyrealsense2` Python library. This library reads the intrinsic parameters of the depth camera's and the RGB module and transforms the depth data to RGB module's reference frame.
%     \item The second step of calibration, is to find the transformation matrix between the RGB Module and the DSLR camera. We use 7x5 ChArUco pattern \ct{cite opencv charuco?} to calibrate between these two cameras. \ct{add notations here for pose transformations, like $T_{rsense-rgb}^{DSLR}$}
%     \ct{Paste a photo of our charuco board as well}
%     \item To ensure that we don't have to recalibrate the sensors everytime, we 3D printed a mount to attach the Depth sensor to the DSLR camera. This is shown in fig x and fig y.
%     \item We capture 15 calibration pictures of ChArUco board in various orientations with both the RGB module and DSLR camera. Then, compute the rotation($R_i$) and translation($t_i$) matrices/vectors for each image pair. We finally compute the average of all the $R_i$ and $t_i$ matrices to get the final and rotation and translation matrix.
%     \item \ct{Mention which opencv function we use}
% \end{itemize}

% \begin{figure}
%     \centering
%     \includegraphics[width=0.5\linewidth]{figures/mount.png}
%     \caption{CAD file of the 3D printed mount to attach the Intel RealSense Depth Camera to the Canon DSLR's hot shoe. We will release the CAD file.}
%     \label{fig:mount}
% \end{figure}

% Write out the calibration procedure. Taking 15 images of the charuco board from both cameras. Then run OpenCV code to estimate the extrinsics and intrinsics for the cameras. 
\section{Per-scene metrics}
\label{sec:sec4}
We report the per-scene metrics in tables~\cref{tab:psmetric1,tab:psmetric2,tab:psmetric3,tab:psmetric4,tab:psmetric5,tab:psmetric6,tab:psmetric7}.
While some competing methods (e.g., Metric3D, MLPro) perform marginally better on a few scenes such as \textsc{thordog} and \textsc{shoerack}, they exhibit significant failures on others.
In contrast, our method demonstrates consistent performance across all scenes, with substantial improvements in the average metrics reported in Table~1 of the main paper.

\begin{table}
\centering
\small
\begin{tabular}{l|cccccc}
\toprule
\textbf{Method} & \textbf{RMSE $\downarrow$} & \textbf{REL $\downarrow$} & \textbf{log10 $\downarrow$} & $\boldsymbol{\delta_1}~\uparrow$ & $\boldsymbol{\delta_2}~\uparrow$ & $\boldsymbol{\delta_3}~\uparrow$ \\
\midrule
Ours       & 0.505 & 0.205 & 0.075 & 0.772 & 0.940 & 0.984 \\
MLPro      & 0.633 & 0.284 & 0.106 & 0.598 & 0.921 & 0.969 \\
UniDepth   & 0.424 & 0.182 & 0.083 & 0.573 & 0.952 & 0.992 \\
Metric3D   & 0.422 & 0.177 & 0.076 & 0.782 & 0.936 & 0.983 \\
\bottomrule
\end{tabular}
\caption{Depth estimation metrics for the \textsc{thordog} scene.}
\label{tab:psmetric1}
\end{table}

\begin{table}
\centering
\small
\begin{tabular}{l|cccccc}
\toprule
\textbf{Method} & \textbf{RMSE $\downarrow$} & \textbf{REL $\downarrow$} & \textbf{log10 $\downarrow$} & $\boldsymbol{\delta_1}~\uparrow$ & $\boldsymbol{\delta_2}~\uparrow$ & $\boldsymbol{\delta_3}~\uparrow$ \\
\midrule
Ours       & 0.312 & 0.157 & 0.067 & 0.805 & 0.968 & 0.985 \\
MLPro      & 0.302 & 0.174 & 0.080 & 0.733 & 0.976 & 0.990 \\
UniDepth   & 1.264 & 0.610 & 0.190 & 0.280 & 0.527 & 0.768 \\
Metric3D   & 0.263 & 0.155 & 0.069 & 0.774 & 0.975 & 0.991 \\
\bottomrule
\end{tabular}
\caption{Depth estimation metrics for the \textsc{books} scene.}
\label{tab:psmetric2}
\end{table}

\begin{table}
\centering
\small
\begin{tabular}{l|cccccc}
\toprule
\textbf{Method} & \textbf{RMSE $\downarrow$} & \textbf{REL $\downarrow$} & \textbf{log10 $\downarrow$} & $\boldsymbol{\delta_1}~\uparrow$ & $\boldsymbol{\delta_2}~\uparrow$ & $\boldsymbol{\delta_3}~\uparrow$ \\
\midrule
Ours       & 0.261 & 0.108 & 0.042 & 0.933 & 0.990 & 0.992 \\
MLPro      & 0.750 & 0.399 & 0.218 & 0.005 & 0.210 & 0.995 \\
UniDepth   & 0.288 & 0.121 & 0.050 & 0.977 & 0.991 & 0.993 \\
Metric3D   & 0.558 & 0.281 & 0.100 & 0.582 & 0.986 & 0.990 \\
\bottomrule
\end{tabular}
\caption{Depth estimation metrics for the \textsc{stairs} scene.}
\label{tab:psmetric3}
\end{table}

\begin{table}
\centering
\small
\begin{tabular}{l|cccccc}
\toprule
\textbf{Method} & \textbf{RMSE $\downarrow$} & \textbf{REL $\downarrow$} & \textbf{log10 $\downarrow$} & $\boldsymbol{\delta_1}~\uparrow$ & $\boldsymbol{\delta_2}~\uparrow$ & $\boldsymbol{\delta_3}~\uparrow$ \\
\midrule
Ours       & 0.064 & 0.049 & 0.022 & 0.996 & 0.998 & 0.999 \\
MLPro      & 0.096 & 0.085 & 0.035 & 0.995 & 0.998 & 0.999 \\
UniDepth   & 1.000 & 0.940 & 0.287 & 0.002 & 0.003 & 0.686 \\
Metric3D   & 1.067 & 1.001 & 0.300 & 0.002 & 0.004 & 0.362 \\
\bottomrule
\end{tabular}
\caption{Depth estimation metrics for the \textsc{plane} scene.}
\label{tab:psmetric4}
\end{table}

\begin{table}
\centering
\small
\begin{tabular}{l|cccccc}
\toprule
\textbf{Method} & \textbf{RMSE $\downarrow$} & \textbf{REL $\downarrow$} & \textbf{log10 $\downarrow$} & $\boldsymbol{\delta_1}~\uparrow$ & $\boldsymbol{\delta_2}~\uparrow$ & $\boldsymbol{\delta_3}~\uparrow$ \\
\midrule
Ours       & 0.346 & 0.125 & 0.058 & 0.827 & 0.951 & 0.990 \\
MLPro      & 0.956 & 0.454 & 0.157 & 0.162 & 0.670 & 0.986 \\
UniDepth   & 0.617 & 0.260 & 0.136 & 0.370 & 0.814 & 0.949 \\
Metric3D   & 0.504 & 0.210 & 0.095 & 0.611 & 0.899 & 0.951 \\
\bottomrule
\end{tabular}
\caption{Depth estimation metrics for the \textsc{toys} scene.}
\label{tab:psmetric5}
\end{table}

\begin{table}
\centering
\small
\begin{tabular}{l|cccccc}
\toprule
\textbf{Method} & \textbf{RMSE $\downarrow$} & \textbf{REL $\downarrow$} & \textbf{log10 $\downarrow$} & $\boldsymbol{\delta_1}~\uparrow$ & $\boldsymbol{\delta_2}~\uparrow$ & $\boldsymbol{\delta_3}~\uparrow$ \\
\midrule
Ours       & 0.251 & 0.121 & 0.053 & 0.903 & 0.996 & 0.998 \\
MLPro      & 0.329 & 0.176 & 0.069 & 0.894 & 0.992 & 0.998 \\
UniDepth   & 0.422 & 0.223 & 0.112 & 0.495 & 0.982 & 0.998 \\
Metric3D   & 0.209 & 0.105 & 0.045 & 0.962 & 0.997 & 0.998 \\
\bottomrule
\end{tabular}
\caption{Depth estimation metrics for the \textsc{shoerack} scene.}
\label{tab:psmetric6}
\end{table}

\begin{table}
\centering
\small
\begin{tabular}{l|cccccc}
\toprule
\textbf{Method} & \textbf{RMSE $\downarrow$} & \textbf{REL $\downarrow$} & \textbf{log10 $\downarrow$} & $\boldsymbol{\delta_1}~\uparrow$ & $\boldsymbol{\delta_2}~\uparrow$ & $\boldsymbol{\delta_3}~\uparrow$ \\
\midrule
Ours       & 0.173 & 0.109 & 0.044 & 0.919 & 0.982 & 0.991 \\
MLPro      & 0.206 & 0.154 & 0.070 & 0.791 & 0.980 & 0.993 \\
UniDepth   & 0.283 & 0.219 & 0.106 & 0.454 & 0.970 & 0.995 \\
Metric3D  & 0.191 & 0.135 & 0.057 & 0.837 & 0.981 & 0.991 \\
\bottomrule
\end{tabular}
\caption{Depth estimation metrics for the \textsc{kitchen} scene.}
\label{tab:psmetric7}
\end{table}

\section{Ablations}
In \cref{tab:zTablation} and \cref{tab:numsteps}, we demonstrate the robustness of our approach to different initializations of $\zT$, and the number of sampling steps (value of $T$). Optimizing for consistency with defocus cues ensures that we obtain consistent metrics that are sufficiently robust (with small standard deviations, see \cref{tab:zTablation}) to the initial noise latents. We observe that while increasing the number of inference steps leads to visually better looking results in some cases, it also increases texture-depth coupling.  
\label{sec:sec5}
\begin{table}[h]
\centering
\resizebox{\textwidth}{!}{%
\begin{tabular}{l|ccccccc}
\toprule
Scene & $\delta_1$ & $\delta_2$ & $\delta_3$ & $\log_{10}$ & REL & RMSE(m) & Runtime (s) \\
\midrule
\textsc{stairs} & 0.87 (0.03) & 0.99 (0.0006) & 0.99 (0.0004) & 0.049 (0.0048) & 0.12 (0.0089) & 0.27 (0.01) & 229.93 \\
\textsc{shoerack} & 0.90 (0.017) & 0.98 (0.01) & 0.99 (0.005) & 0.052 (0.0036) & 0.12 (0.0086) & 0.25 (0.02) & 229.91 \\
\bottomrule
\end{tabular}
}
\caption{Mean (std. dev) for depth metrics from our method over 10 runs initialized with different noise latents $\zT$ for the \textsc{stairs} and \textsc{shoerack} scenes. We observe considerable robustness to initial $\zT$ initialization as demonstrated by low standard deviations.}
\label{tab:zTablation}
\end{table}

\begin{table}[h]
\centering
\begin{tabular}{l|ccccccc}
\toprule
Sampling steps & $\delta_1$ & $\delta_2$ & $\delta_3$ & $\log_{10}$ & REL & RMSE (m) & Runtime (s) \\
\midrule
1 & 0.8971 & 0.9807 & 0.9926 & 0.0477 & 0.1114 & 0.2345 & 228.07 \\
2     & 0.8739 & 0.9799 & 0.9926 & 0.0500 & 0.1150 & 0.2352 & 328.68 \\
3     & 0.8578 & 0.9666 & 0.9933 & 0.0569 & 0.1293 & 0.2501 & 430.27 \\
4     & 0.8913 & 0.9796 & 0.9928 & 0.0514 & 0.1217 & 0.2410 & 530.25 \\
\bottomrule
\end{tabular}
\caption{Varying the number of sampling steps in Marigold-LCM. We observe that increasing the number of sampling steps leads to higher optimization runtime, with only marginal improvements in quantitative metrics. We thus opt for a single sampling step in our method.}
\label{tab:numsteps}
\end{table}

% \begin{itemize}
%     \item  Effect of different zT initialization 
%     \item Number of sampling steps. 
% \end{itemize}

\section{Visualizing Marigold predictions pre and post optimization}
\label{sec:sec6}
\begin{figure}
    \centering
    \includegraphics[width=\linewidth]{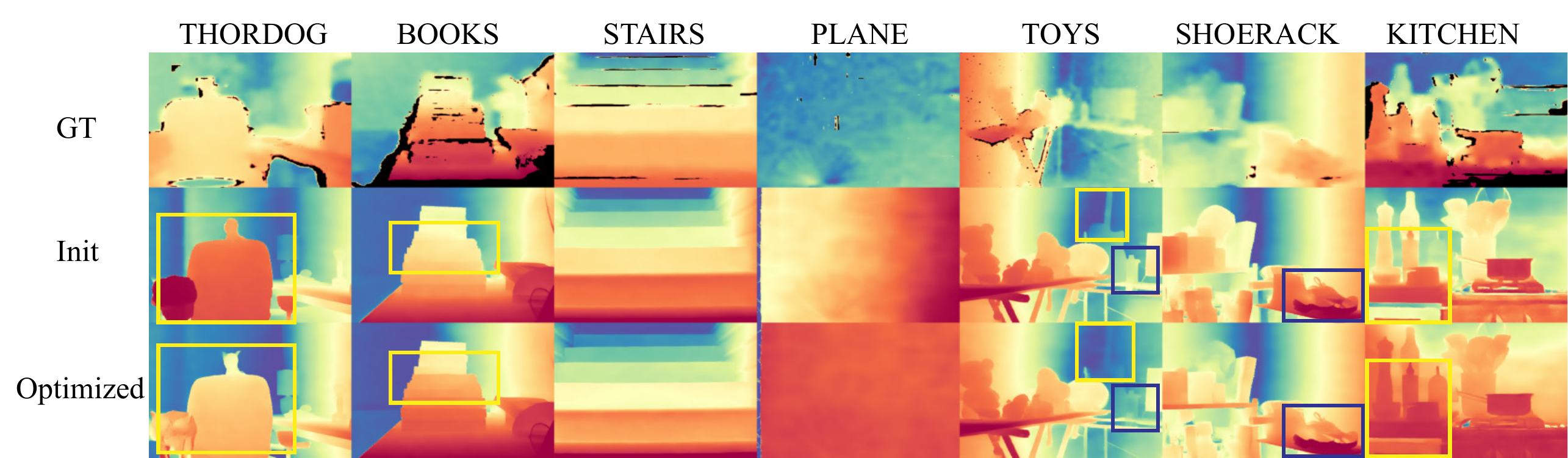}
    \caption{Marigold relative depth after optimization. We show regions (yellow boxes) where we observe improvements from our method (bottom row) compared to the initial relative depth prediction (middle row) from Marigold. We observe failure cases (blue boxes) in the \textsc{shoerack} scene, where the optimization fails to correct for the wrong initial prediction by Marigold. In the \textsc{toys} scene, while our model reduces the detail on the bottle (blue box), it recovers the correct relative depth region for the guitar region (yellow box).}
    \label{fig:relcorrection}
\end{figure}

We quantitatively showed in Table 2 of the main paper that optimizing $\zT$, i.e., refining both the relative depth and affine parameters, yields better results than optimizing only the affine scaling while keeping the Marigold-predicted depth fixed. In \cref{fig:relcorrection}, we visualize how incorporating defocus cues affects the relative depth predicted by Marigold.

In some regions of the \textsc{shoerack} scene (blue boxes/insets), our method struggles to correct a poor initial estimate for one of the shoes. However, across most other scenes (\textsc{thordog, plane, kitchen, books}), it visibly improves relative depth in several regions (see yellow insets). Recall that the final metric depth output depends on both the refined relative depth and the learned affine parameters. For example, in the \textsc{kitchen} scene, our method sharpens relative depth for objects like the bottles and stove (yellow inset), and compensates for upper-scene errors by adjusting the affine offset $\alpha$, leading to a more accurate metric depth reconstruction (see Fig. 4 in the main paper).
\begin{table}
\centering
\begin{tabular}{l|cccccc}
\toprule
F-stop ($N$) & $\delta_1$ & $\delta_2$ & $\delta_3$ & $\log_{10}$ & REL & RMSE (m) \\
\midrule
4.0  & 0.8578 & 0.9499 & 0.9747 & 0.0536 & 0.1216 & 0.2722 \\
8.0  & 0.8971 & 0.9807 & 0.9926 & 0.0477 & 0.1114 & 0.2345 \\
11.0 & 0.8971 & 0.9802 & 0.9929 & 0.0507 & 0.1164 & 0.2335 \\
13.0 & 0.6571 & 0.9088 & 0.9858 & 0.0905 & 0.1912 & 0.3682 \\
16.0 & 0.5472 & 0.8539 & 0.9884 & 0.1107 & 0.2226 & 0.4172 \\
\bottomrule
\end{tabular}
\caption{Depth estimation performance across different F-stop values. We report numbers averaged over all scenes, excluding \textsc{thordog} since it doesn't have images for $N=11, 16$. We observe that $N=8$ is optimal across most metrics, with performance degrading for higher or lower F stops. We thus report results with $N=8$ in the main paper. }
\label{tab:diffapertures}
\end{table}

\section{Performance across different apertures on real data}
\label{sec:sec7}
Recall that in Fig. 5 (Left) of the main paper, we evaluate our method on blurred images simulated at different apertures. This synthetic experiment allows us to simulate a wider range of apertures for comparison, while isolating the effect of forward model mismatch from the effect of aperture size. In \cref{tab:diffapertures}, we show that the trend seen in Fig. 5 main paper -- optimal performance at a specific F-stop with reduced accuracy at larger or smaller apertures, holds true for real-world data as well. We obtain optimal performance at $N=8$ on the real dataset, with degraded results at both smaller and larger F-stops. The optimal F-stop may differ between real and synthetic datasets, as performance in real data is influenced by both forward model mismatch (which depends on aperture size) and the strength of the blur cue.

\end{document}